\documentclass[11pt]{article}

\usepackage[final]{acl}

\usepackage{times}
\usepackage{latexsym}
\usepackage{subcaption}
\usepackage{booktabs}
\usepackage[table]{xcolor}
\usepackage{float}
\definecolor{lightbluehl}{RGB}{225,240,255}
\usepackage{listings}
\usepackage{xcolor}
\usepackage{comment}
\definecolor{codegray}{RGB}{248,248,248}
\definecolor{lightyellowhl}{RGB}{255, 249, 196}

\lstdefinestyle{promptstyle}{
  basicstyle=\ttfamily\footnotesize,
  backgroundcolor=\color{codegray},
  frame=single,
  breaklines=true,
  breakatwhitespace=true,
  columns=fullflexible,
  keepspaces=true,
  showstringspaces=false,
  xleftmargin=0.5em,
  xrightmargin=0.5em,
  aboveskip=0.8em,
  belowskip=0.8em
}

\usepackage[T1]{fontenc}

\usepackage[utf8]{inputenc}

\usepackage{microtype}

\usepackage{inconsolata}

\usepackage{graphicx}
\usepackage{booktabs}
\usepackage{multirow}
\usepackage{amsmath}
\usepackage{graphicx}
\usepackage{caption}
\title{\textsc{Empath}: Tracing Multi-Level Emotion Dynamics in Crisis Counseling Dialogues }

\author{
    \textbf{Ziwei Gong$^{1^\ast}$}, 
    \textbf{Yuchen Huang$^{2^\ast}$}, 
    \textbf{Wen Liang$^1$}, 
    \textbf{Nicholas Deas$^1$},
    \textbf{Melanie Subbiah$^1$},\\
    \textbf{Kathleen McKeown$^1$},
    \textbf{Julia Hirschberg$^1$},
    \\
    $^1$Columbia University, USA~~~~~~
    $^2$Barnard College, USA
    \\
    \small{\texttt{\{sara.ziweigong, ndeas, kathy, julia\}@cs.columbia.edu}}
   \\
   \small{$^\ast$Equal contributions.}
}

\newcommand{\sara}[1]{\textcolor{pink}{sara: #1}}
\newcommand{\ndnote}[1]{\textcolor{teal}{ND: #1}}
\newcommand{\wen}[1]{\textcolor{blue}{#1}}
\newcommand{\lareina}[1]{\textcolor{green}{#1}}
\newcommand{\msnote}[1]{\textcolor{orange}{MS: #1}}

\renewcommand{\sara}[1]{}
\renewcommand{\ndnote}[1]{}
\renewcommand{\wen}[1]{}
\renewcommand{\lareina}[1]{}
\renewcommand{\msnote}[1]{}
\newcommand{\sararevision}[1]{\textcolor{magenta}{Sara: #1}}

\newcommand{\empath}{\textsc{Empath}}

\begin{document}
\maketitle
\begin{abstract}
% This document is a draft for EMNLP paper - synthetic data comparison through emotion distributions and transitions. 

Emotion dynamics are critical for understanding crisis-support conversations, yet most computational work treats emotion as static utterance-level labels. We introduce \empath{}, a framework for understanding affective dynamics in mental health dialogues across three granularities: turn-level labels, transition probabilities, and global conversation archetypes. Applying \empath{} to text-based crisis conversations with self-identified Black texters discussing grief, we find persistent negative affect, gradual hope-ward transitions, distinct texter–volunteer emotional roles, and heterogeneous recovery trajectories. 
% \kmnote{Could possibly remove this sentence and just put in intro. }
% A secondary comparison with synthetic crisis dialogues shows that synthetic data captures some aggregate trends but misses fine-grained transition and archetypal diversity. 
These results highlight the informative patterns that emerge from computationally understanding crisis support and expressions of grief as dynamic processes within conversations, as well as the overall value of emotion-dynamic analysis for analyzing and comparing affect in dialogues.
% and highlight the value of emotion-dynamic analysis for studying authentic support and evaluating privacy-preserving substitutes. 
\end{abstract}

\section{Introduction}

Understanding how emotions change over the course of crisis-support conversations is critical for studying online grief support. In text-based crisis settings, texters may move through persistent distress, disclosure, moments of gratitude, and gradual shifts toward hope over the course of a single interaction. These changes are especially important in counseling conversations, where support is rarely a simple linear movement from negative to positive emotion. However, much computational work on mental health dialogue treats emotion as a static utterance-level label, following broader trends in emotion recognition and affective state identification \cite{jordan2025speech, wu-etal-2024-multimodal, gong-etal-2023-eliciting}, leaving the temporal structure of emotional change underexplored. 
Related work also examines the extraction of texters' explicit emotion expressions in crisis conversations \cite{buda-etal-2026-extraction}.

% As part of a larger collaboration with social work researchers leveraging NLP tools to better understand grief, we examine conversations held on Crisis Text Line (CTL), a mental health organization where volunteer counselors provide support over text messages to those in crisis, as a benchmark for authentic crisis support. \ndnote{expand on this} \sara{expaned}
In this work, we study emotion dynamics in conversations from Crisis Text Line (CTL),~\footnote{\url{https://www.crisistextline.org}} a mental health organization where volunteer counselors provide support over text messages to those in crisis.
% , as a benchmark for authentic crisis support. 
As part of a larger collaboration between computational linguists and social work researchers, we focus specifically on CTL conversations with Black texters discussing or expressing grief.
Prior work has emphasized that grief in Black communities is shaped by social, historical, and structural contexts \cite{10.3389/fpsyt.2022.850994}, and emotional expressions of grief are known to be highly complex, sometimes involving simultaneous expressions of joy and deep distress \cite{patton-grief}. 
% motivating analyses that attend to both emotional expression and support-seeking practices \cite{10.3389/fpsyt.2022.850994}. 
This setting is therefore a challenging domain for analyzing highly complex emotion expression, and provides an opportunity to examine how distress and support unfold in 
% authentic 
crisis conversations.
This setting also raises important methodological challenges: the data is highly sensitive, privacy-restricted, and cannot be processed using API-based systems or publicly released at scale. 
% \kmnote{The fact that these are Black texters seems gratuitous both here and in contributions list. Are you able to strengthen by any observers about shifts or emotions that are particularly interesting given the demographics? I think TCL would look for this also. You could do this in a qualitative study. }
% Similar privacy and access constraints have motivated both domain-specific mental health NLP models and privacy-preserving alternatives for clinical dialogue analysis \cite{ji-etal-2022-mentalbert,cabrera-lozoya-etal-2025-synthetic}.

% \ndnote{Instead of synthetic alternatives below, just introduce as a tool to analyze conversations}
% To evaluate the fidelity of synthetic alternatives, we propose the \empath{} framework, which identify emotion and analyze conversation in 3 levels: emotion distributions, measuring aggregate affective prevalence; micro-dynamics, tracking turn-level transitions and polarity shifts; and macro-dynamics, characterizing counseling strategies and conversation archetypes.
% \ndnote{layout framework a bit more} \sara{done}

To analyze these conversations, we propose \empath{}, a framework for studying emotion dynamics in therapeutic and crisis-support dialogue across three levels of granularity. First, \empath{} identifies utterance-level emotions using a locally runnable emotion-labeling pipeline designed for privacy-restricted data. Second, it analyzes micro-dynamics, including turn-level emotion transitions, polarity shifts, persistence, and recovery pivots. 
Third, it characterizes macro-dynamics, including volunteer strategies and conversation-level trajectory archetypes. Together, these levels allow us to move beyond aggregate emotion distributions and examine how emotional states persist, shift, and resolve over interactions.
% We first analyze the emotion dynamics within CTL conversations, serving as authentic mental health-centered dialogues. We them evaluate and compare synthetic versus authentic dialogues at three levels that reflect the task’s goals: (1) emotion distributions, the overall prevalence and proportion of affective states across the dataset; (2) emotion transitions and polarity, the dynamic, turn-by-turn trajectory of emotional shifts and de-escalation over time; (3) conversation macro-dynamics, looking into conversation strategies and archetypes.
% This three-level evaluation framework processes raw dialogue through an analysis ranging from individual utterances to holistic conversation: \textit{i.} Utterance-Level Emotion Detection; \textit{ii.}  Micro-Dynamics (Transitions and Polarity); and \textit{iii.}  Macro-Dynamics (Strategies and Archetypes).

% \ndnote{Summarize application to CTL, briefly introduce and motivate other domains} \sara{updated - but maybe we can select the most interesting findings? currently I'm pulling findings I remembered}

Applying \empath{} to CTL grief conversations, we 
% find that 
% authentic 
surface a range of different emotion dynamics patterns, including repeated negative emotion expressions, shifts from negative emotions to hope, and mixtures of these patterns. Such emotion dynamics also highlight  distinctions between the emotional roles of texters and volunteer counselors in dialogues.
% crisis support is characterized by persistent negative affect, gradual hope-ward transitions, distinct emotional roles for texters and volunteer counselors, and heterogeneous recovery trajectories.
% \kmnote{Repeats abstract with no new information. Could you have an example of at least one of these but preferably more? Another possibility would be to include more of a definition? Some of these things seem unclear.}
Examining these patterns suggests that successful crisis support is better understood as a dynamic process rather than a terminal shift from distress to resolution. In particular, the timing and structure of emotional transitions reveal aspects of support that are not visible from traditional utterance-level labels or conversation endpoints alone.

Finally, we use synthetic crisis-support dialogues to show how the framework also applies beyond CTL.
% as a secondary comparison.
Synthetic data has become increasingly attractive for developing analysis tools while preserving the confidentiality of real conversations \cite{kurakin-privacy, flemings-privacy,cabrera-synthetic}, and recent work has explored LLM-based patient simulations for clinical and counselor training \cite{louie-etal-2024-roleplay, Louie_2026, wang-etal-2024-patient}. 
% However, it remains unclear whether synthetic conversations preserve the affective and interactional structure of authentic crisis support. 
We show that \empath{} can reveal meaningful differences in emotion dynamics, surfacing some similarities in aggregate emotional trends but gaps in fine-grained patterns between synthetic and authentic dialogues.

We summarize our contributions as follows:
\begin{enumerate}
    \setlength{\itemsep}{0pt}
    \setlength{\parskip}{0pt}
    \setlength{\parsep}{0pt}
    \item We propose a \textbf{novel, three-level evaluation framework, \empath{}, to assess emotion dynamics} in mental health and crisis dialogues.
    \item Using \empath{}, we conduct a \textbf{large-scale analysis of text-based crisis conversations with Black texters about grief} from Crisis Text Line. We discuss how trends in emotion dynamics are reflective of successful dialogues in this context. We identify key dynamic signatures of grief-support conversations, including persistent distress, hope-ward pivots, role-differentiated emotional profiles, and heterogeneous recovery trajectories.
    % \item We explore how the \empath{} framework extends to two additional contexts: synthetically-generated conversations. We find that synthetic data can match some endpoint patterns while missing important transition-level and trajectory-level structure. 
    % \begin{comment}
    % \item We \textbf{further validate \empath{} on another type of dialogue}--synthetic conversations--showing how the framework reveals important gaps in transition-level and trajectory-level structures.
    % \end{comment}
    % \ndnote{Probably need some kind of takeaway here.} \sara{added takeaway for both, but we can downplay the last point too}\ndnote{Rewrote to downplay}
    
\end{enumerate}

\section{Related Work}

\textbf{Emotion Recognition in Conversation} (ERC) is a foundational NLP task and key tool for tracking crisis intervention trajectories \cite{tripodi-etal-2025-assessing, Xu_2024}. In text-based therapeutic dialogues, model evaluations reveal clear tradeoffs: closed-source models like GPT-4 perform well in zero-shot diagnostic settings but often struggle with nuanced, culturally sensitive emotional states \cite{wu-etal-2025-multimodal, xie2025culturalpromptingimprovesempathy}. Open-weight LLaMA models, when lightly fine-tuned, achieve competitive performance in assessing emotional safety \cite{badawi-etal-2026-trust}. Domain-specific models, such as MentalBERT and MentalRoBERTa \cite{ji2022mentalbert}, consistently outperform general models on targeted clinical tasks, while crisis-focused models like BERT-EV leverage continuous valence scoring to track turn-level de-escalation in real-world crisis conversations \cite{tripodi-etal-2025-assessing}. Efforts on crisis de-escalation and emotional support dialogue suggest that interactional trajectories, support strategies, and changes in distress over time are central to understanding support effectiveness
\cite{tripodi-etal-2025-assessing,liu-etal-2021-towards, wan-etal-2025-emodynamix, zhang-etal-2025-intentionesc, liu-etal-2026-review}. 
% \ndnote{Do we nede any more stuff here?}

\textbf{Grief and bereavement} are highly complex experiences that are often misunderstood \cite{hall-grief} and accompanied by constantly evolving theories including dual process models \cite{stroebe-dual} and meaning making-focused perspectives \cite{stroebe-meaning, neimeyer-meaning}. Understanding these experiences has been further complicated by the movement of their expression to online, digital spaces \cite{moore-digital, patton-grief}. 
% Our work is situated within a larger effort to study experiences and digital expression of grief by members of the Black community. 
As expressions of grief in digital counseling are highly dynamic and complex, we focus on this domain as a potential case where understanding may be aided through emotion dynamics.

\begin{figure*}[htbp]
    \centering
    \includegraphics[width=.9\linewidth]{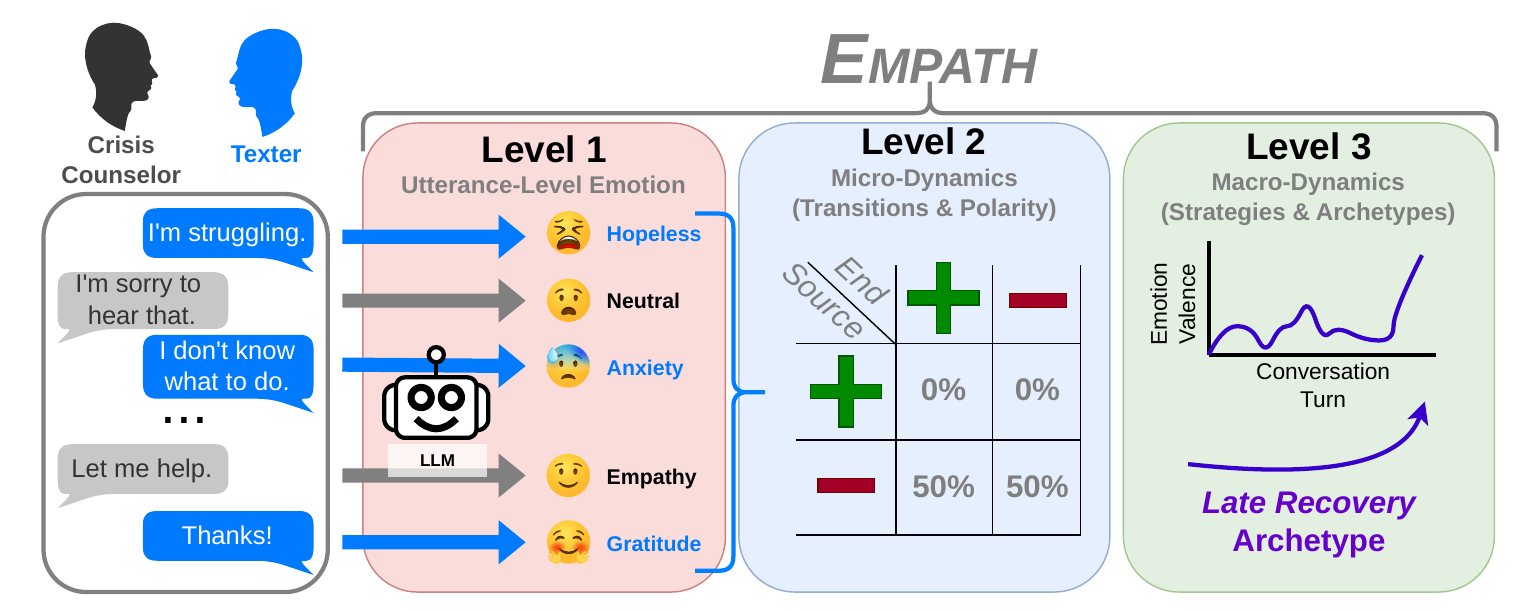}
    \caption{Summary of \empath{} framework for assessing conversation dynamics in mental health/crisis dialogues. The example dialogue is simulated and contains no real CTL messages. 
    % \ndnote{Made this just a general framework diagram instead of specifically for synthetic data. \sara{love the new figure!}}
    % \msnote{I would clean this up more but putting here as a starting point to discuss what needs to go in this figure.}\ndnote{Only things I might change is to just call the LLM box in the middle like emotion classifier since any emotion analysis method could be used here. Otherwise, I like it!}\lareina{do we also want to add the zero-shot? --sara:yes}\ndnote{Showing both setups on the left side might be good}
    }
    \label{fig:pipeline}
\end{figure*}

% \section{Datasets}
\section{Crisis Dialogue Data}
\label{sec:crisis_data}
% The Crisis Text Line (CTL) dataset consists of anonymized text-based crisis counseling conversations between texters in distress and trained volunteer counselors \cite{althoff2016counseling}. The dataset used in this study contains \textbf{2,478 conversations} comprising \textbf{133,666 total utterances} from three speaker types: texters (62,723 utterances), volunteer counselors (56,460 utterances), and system messages (14,483 utterances). Conversations span a range of crisis topics including depression, anxiety, suicidal ideation, relationship issues, and self-harm.

Crisis Text Line (CTL) is a non-profit mental health organization that provides real-time support for those in crisis over text messages. Those in crisis (\textit{Texters}) that reach out to CTL are connected with a volunteer crisis counselor (\textit{Volunteers}) to receive support. All crisis counselors are trained by CTL to provide effective support to texters and are supervised live by trained mental health professional staff.

% \msnote{I think it would be helpful having another sentence contextualizing the focus on Black texters, since it's not mentioned anywhere in the focus of this study. You could also reduce the emphasis on this point and just include a shorter sentence saying that to reduce coding load on the social work team, you overlap with their study and focus on Black texters.} \ndnote{Nick to-do}

We use a corpus of 2,478 de-identified conversations focused on expressions of grief in the Black community. We collect only conversations with texters that self-identified as Black or African American in an optional post-conversation survey. CTL volunteers may tag conversations with a topic, and we collect only conversations tagged as discussing \textit{grief}. 
% through a signed Data Use Agreement (DUA) with CTL, 
From this corpus, we collect a random sample of 100 de-identified conversations for human annotation and validation of the emotion models. Importantly, due to the sensitive nature of the data and in order to protect the confidentiality of CTL users, this data is de-identified prior to access, is only accessed under a signed DUA, and all data is processed locally (i.e., no CTL data are provided to API-based LLMs or other online services). 
% \sararevision{Do we need to add the actual IRB determination and protocol number?}
% \ndnote{Lay this out a bit more}

We collaborate with a team of social work researchers who conduct an inductive thematic analysis of the conversations--the researchers thoroughly read and label conversation turns according to themes that are qualitatively derived from the data. All conversations within the 100-conversation annotation sample are annotated by two researchers, and all disagreements between them are resolved through discussion. 
% These themes cover a variety of categories, including events that caused the texters' crisis (e.g., loss of family member), discussion of coping strategies (e.g., journaling/writing), and mention of symptoms that often co-occur with depression (e.g., loss of appetite). In this work, 
While these themes cover a variety of categories, in this work, we focus on a set of 37 emotions (e.g., \textit{hopeful}, \textit{numbness}, \textit{longing}) identified by the social work researchers, as defined in Table~\ref{tab:emotion_labels}.
% . A full list of the emotions are included in Table~\ref{tab:emotion_labels}.
Text-level statistics summarizing the annotation sample are included in Table~\ref{tab:data-stats}. The 100 conversations span a period from December 2016 to October 2023, with an average of 56.3 dialogue turns each.

% \sararevision{Confirm the definition of dialogue turns and the reported average against Table~\ref{tab:data-stats}; retain the original value until verified.}

% \ndnote{There was a short paragraph here about the synthetic conversations. Making a note in case we need to bring this back in later.}
% We additionally construct a synthetic crisis-counseling corpus to compare and contrast with authentic CTL conversations. We describe the full generation setup in \S\ref{sec:generation_pipeline}. 

% \subsection{Synthetic Data Description}
% \msnote{This section is very short and the content will be elaborated on later. I think you could remove it and focus exclusively on the real data in this section} We construct a synthetic crisis-counseling corpus to compare and contrast with authentic CTL conversations. Our synthetic conversations are generated with LLM-based role simulation, where prompted agents produce texter--volunteer exchanges under controlled topic seeds and conversation constraints. We describe the full generation setup, including the dual-agent and zero-shot variants, in \S\ref{sec:generation_pipeline}.

\begin{table}[t]
    \centering
    \resizebox{0.8\linewidth}{!}{%
    \begin{tabular}{cccc}
        \toprule
         Author & Count & \# Labeled & Avg. Len \\
         \midrule
         System     & 629   & 0     & 24.2 \\
         Texter     & 3355  & 474   & 17.7 \\
         Volunteer  & 2695  & 19    & 25.2 \\
         \midrule
         Overall    & 6679  & 493   & 21.4 \\
         \bottomrule
    \end{tabular}
    }
    \caption{Message summary statistics of the 100-conversation CTL annotation sample.}
    \label{tab:data-stats}
\end{table}

\section{\empath{} Framework} %Data Generation and Analysis Pipeline
% Our methodology consists of two primary components: 1)~a synthetic data generation pipeline used to produce controlled crisis-counseling dialogues, and 2)~the Emotion \& Multidimensional Pattern Analysis for Therapeutic Help (\empath{}) framework, a multi-level analytic engine used to compare synthetic outputs against real-world benchmarks (CTL).
%\sara{or "ADAPT" (Affective Dynamics & Archetypal Pattern Tracking)?}
% While the generation module is specific to our synthetic corpus, the \textbf{EDC} framework is applied identically to both the CTL and synthetic data to ensure a controlled comparative analysis.
Our primary methodology is the \underline{\textbf{E}}motion and \underline{\textbf{M}}ultidimensional \underline{\textbf{P}}attern \underline{\textbf{A}}nalysis for \underline{\textbf{T}}herapeutic \underline{\textbf{H}}elp (\empath{}) framework. We introduce \empath{} as a tool for studying 
% trends and 
patterns in the emotion dynamics of crisis, mental health, and other therapeutic dialogues at multiple levels. 
%
% We develop a three-stage pipeline: (1)~a synthetic data generation module that produces controlled crisis-counseling conversations, (2)~an LLM-based emotion detection module that assigns one of 37 emotion labels to every utterance, and (3)~an emotion analysis module that characterises the temporal dynamics of those labels across conversations. The latter two stages are applied identically to the CTL and synthetic corpora to enable a controlled comparison.
% \sara{Do we want to name the pipeline?} \msnote{I don't think we should name the data generation pipeline if we're not advocating for people to actually use it for synthetic data, but we could name the emotion evaluation framework.} \sara{I seperated the generation and the emotion as two parts, hopefully this make thing more clear?}
%
% \msnote{first pass in Figure \ref{fig:pipeline}}
%
% \subsection{The \empath{} Framework}
% The \textbf{\empath{}} framework serves as our primary comparative tool, consisting of an LLM-based emotion detection module and modules for micro- and macro-affective conversation dynamics. 
This three-level evaluation framework (summarized in Figure~\ref{fig:pipeline}) processes raw dialogue through an analysis ranging from individual utterances to holistic conversation: \textit{1)}~\textbf{Utterance-Level Emotion Detection}; \textit{2)}~\textbf{Micro-Dynamics} (Transitions and Polarity); and \textit{iii)}~\textbf{Macro-Dynamics} (Strategies, Archetypes, and Diversity). 
% \ndnote{More cuts commented out below}
% The entire framework is applied identically to the CTL and synthetic corpora to enable a controlled comparison. Both stages are applied identically to the CTL and synthetic corpora to enable a controlled comparison. 
% \ndnote{(Nick to-do) Double check we are consistent about "authentic" vs. "CTL"}

% comprises the primary analytic engine of this study, a two-stage pipeline: (1)~an LLM-based emotion detection module that assigns one of 37 emotion labels to every utterance, and (2)~an emotion analysis module that characterises the temporal dynamics of those labels across conversations. Both stages are applied identically to the CTL and synthetic corpora to enable a controlled comparison.

\subsection{Level 1: Emotion Detection}
\label{sec:emotion_detection}

    \begin{table*}[ht]
    \centering
    \small
    \resizebox{\linewidth}{!}{
    \begin{tabular}{p{0.03\linewidth} p{0.18\linewidth} p{0.75\linewidth}}
    \toprule
    \textbf{Level} & \textbf{Statistic} & \textbf{Description} \\
    \midrule
    C & Transition matrix & How often each emotion follows another ($37 \times 37$ counts and probabilities). \\
    C & Persistence & How likely an emotion is to repeat on the next turn ($P(\text{stay})$). \\
    C & Net flow & Whether an emotion is a ``sink'' (conversations flow in) or ``source'' (conversations flow out). \\
    C & Motifs & Common 2- and 3-emotion sequences (e.g., hopeless$\to$hopeless$\to$hopeful). \\
    C & Stationary distribution & Long-run equilibrium prevalence of each emotion under a Markov model. \\
    C & Change rate & Fraction of consecutive turns where the emotion label changes. \\
    P & Major flips & Direct negative$\to$positive or positive$\to$negative sentiment shifts, with timing of the first flip. \\
    P & Within-polarity shifts & Emotion changes that stay within the same sentiment (e.g., sadness$\to$anxiety, both negative). \\
    P & Cross-polarity shifts & Which specific emotions are involved at each sentiment boundary crossing. \\
    Both & Sankey diagrams & Visual flow diagrams showing how emotions redistribute from conversation start to end. \\
    \bottomrule
    \end{tabular}
    }
    %\msnote{I find these descriptions a bit confusing} \wen{Simplified all descriptions to plain language.}
    \caption{Micro-Dynamics Statistics derived from emotion-label sequences at the category level of 37 emotion labels (C) and polarity level of positive/negative/neutral (P).}
    \label{tab:analysis_metrics}
    \end{table*}
    
    % \subsubsection{Model Design}
    
    We frame utterance-level emotion detection as a constrained text generation task, similar to recent emotion analysis work \cite{deas-masive}. Given an utterance and optional conversational context, a language model is prompted to select exactly one label from a list of the 37 emotion categories (Table~\ref{tab:emotion_labels}) and provide reasoning.
These labels describe emotional content expressed or reflected in an utterance; volunteer labels may reflect texters' emotions rather than the volunteers' own emotional states.
    
    \paragraph{Prompt Design.}
    Each prompt consists of three components: \textit{i)}~the list of 37 emotion labels with brief
    % natural-language definitions, 
    descriptions, \textit{ii)}~the conversational history (a sliding window of up to $N$ preceding utterances from both speakers), and \textit{iii)}~an instruction asking the model to output a structured \textit{(label, reason)} response with post-hoc explanations \cite{camburu-explain,limpijankit-counterfactual}. 
    % \ndnote{(Nick to-do) Might be a little odd to reviewers that reasoning comes after, cite some of the simulatability work to motivate this.}
    The instruction explicitly directs the model to weight the current message most heavily and to use preceding messages only as supporting context. 
    % We design a prediction strategy with fallback options to ensure valid outputs. 
    % We evaluate two inference modes: without added context (target utterance only) and with context (target plus $N$ preceding utterances). 
    Prompts provided in \S\ref{sec:prompt_appendix}.
    
    \paragraph{Model Selection and Validation.} 
    % Since the data-sharing agreements for CTL data strictly prohibit the use of cloud-based commercial APIs, we 
    To safeguard the confidentiality of the CTL conversations, 
    % and ensure adherance to data privacy protocols, 
    % we avoid commercial APIs and 
    we exclusively evaluate locally runnable models within a secure, offline environment.
    % to ensure adherence to data privacy protocols. 
    Specifically, we use \texttt{meta-llama/Llama-3.2-3B-Instruct} \cite{llama3} as the backbone model, with experimental setups detailed in \S\ref{app:setups}.
    We validate the pipeline by comparing generative models (Llama-3.1/3.2, Mistral-7B), encoder-only models (BERT Emotions, MentalBERT), and a random baseline across the human-annotated subset of the CTL conversations.
    % six human-annotated subsets of the CTL corpus. 
    Performance is measured using exact-match accuracy and semantic similarity.

    Our results demonstrate that generative models significantly outperform BERT-based models across all taxonomy levels.
    % , from the fine-grained 37-label set to coarse-grained sentiment categories. 
    % While Mistral-7B achieved the highest raw accuracy, w
    We selected Llama-3.2-3B-Instruct for all primary analyses as it offered the best trade-off between performance and computational efficiency. Notably, it achieved semantic similarity scores (0.97--0.98) comparable to larger models, indicating that its predictions are 
    semantically aligned with human references even when exact labels differ. Full candidate model descriptions, validation results, and the multi-level taxonomy mapping are detailed in \S\ref{app:model_val_appendix}.
    
    %\msnote{I think this is an interesting point and I would prefer to have a little more context in the main body of the paper explaining what you mean by this} \wen{Expanded with specific numbers.}
    % \ndnote{(Nick to-do) Below is repeated from earlier, remove one}
    We also test two configurations per corpus: (i)~\emph{without context} (single-utterance), and (ii)~\emph{with context} 
    % using a sliding window of the 5~most recent 
    (includes preceding utterances). We select (i)~\emph{with context} for our main analysis. 
    % \ndnote{(Nick to-do) Since the evaluation criteria haven't really been introduced yet, can probably move the stats below to the Appendix F and summarize to save space.}
    While context-free models score slightly higher on single-utterance accuracy, the with-context mode produces the temporally coherent trajectories necessary for studying conversation dynamics (see further discussion and full comparison in \S\ref{sec:context_comparison_appendix}).
    % for the full comparison.

    % \ndnote{We should probably make sure we are consistent about using "level", some places say "module" and might be confusing.} \sara{fixed}
    \subsection{Level 2: Micro-Dynamics (Transitions and Polarity)}
    Once labels are assigned, this level characterizes the dynamics of emotional states across conversations
    % The analysis module uses per-utterance labels from the detection stage and characterizes emotion dynamics
    at two complementary levels of granularity: \emph{category-level}, and \emph{polarity-level}. Table~\ref{tab:analysis_metrics} summarizes the full set of derived statistics.
    \textbf{Category-level analysis} tracks transitions among all 37 emotion labels. Each conversation is modeled as a sequence of emotion labels from which we compute adjacent-pair transitions.
    \textbf{Polarity-level analysis} projects labels onto a polarity scale to capture 
    % broad
    coarse sentiment trajectories. 
    To complement the fine-grained emotion category view, we project each of the 37 emotion labels onto a continuous valence axis using the NRC Valence--Arousal--Dominance (VAD) Lexicon~\cite{mohammad2018obtaining} and discretize into three polarity classes: \textit{positive} (valence $\geq 0.55$); \textit{negative} (valence $\leq 0.45$); and \textit{neutral} ($0.45 < \text{valence} < 0.55$).
    % The symmetric 0.05-point margins around the midpoint (0.50) ensure that only clearly valenced labels receive a positive or negative assignment.
    % ; of the 37 labels, 21 map to negative, 12 to positive, and 3 (\texttt{neutral}, \texttt{mood}, \texttt{resilient}) to neutral. 
    The full mapping is provided in \S\ref{sec:polarity_mapping_appendix}.
    % \ndnote{For later, do we need to use "polarity"? I can see people getting confused thinking that polarity is different from valence later in the paper, while most of those that would read this paper would know what valence means pretty well. } \sara{I agree that naming it as emotion/sentiment might make more sense, we can rename later.}

    % Over the resulting three-class polarity sequences we compute the same suite of transition statistics, supplemented by:
    % \begin{itemize}
    %     \item \emph{Major flip} detection: direct negative$\to$positive or positive$\to$negative transitions, with first-flip timing statistics per conversation.
    %     \item \emph{Within-polarity shifts}: emotion-label changes that remain within the same polarity class (e.g., sadness$\to$anxiety, both negative).
    %     \item \emph{Cross-polarity shifts}: detailed label-level information at every polarity boundary crossing.
    % \end{itemize}
    
    % This dual-resolution approach (37-category + 3-polarity) enables comparisons at both the granular emotional level and the high-level sentiment trajectory level.
    % In addition to tabular summaries, we generate Sankey flow diagrams per corpus$\times$author combination: overall, start$\to$end, turn~0$\to$1, turn~1$\to$2, 3rd-to-last$\to$2nd-to-last, and 2nd-to-last$\to$last. 
    
    \subsection{Level 3: Macro-Dynamics (Strategies and Archetypes)}
    Moving beyond surface-level labels, we analyze the interactional structure and global composition of conversations. To support this, we derive a per-turn distress score $d$ from negated valence score  $v$ , $d = 1 - v \in [0, 1]$, so that negative-valence labels (e.g., \textit{fear}, \textit{sadness}) yield high distress values and positive-valence labels (e.g., \textit{gratitude}, \textit{joy}) yield low ones.
    
    \paragraph{Volunteer Strategy Analysis. }

    %\sara{Should we refine the "Strategy Analysis" section to more explicitly state how the distress score $d$ is derived from the 37 emotion labels/VAD mentioned in Level 1? \lareina{added in the opening of 4.2.3 as the distress score is used for all downstream analysis}}
    
    Surface-level emotional arcs may look similar across real and synthetic conversations while masking differences in \emph{how} support is delivered.
    To probe this, we examine whether the same volunteer strategy produces comparable downstream effects on texter distress in synthetic and real CTL data. We analyze local three-turn windows $(u_t, h_t, u_{t+1})$, where a texter turn $u_t$ is followed by a volunteer turn $h_t$ and the next texter turn $u_{t+1}$. Each volunteer turn is classified into one of eight ESConv support-strategy categories \citep{bai2025emotionalsupportersusemultiple}: \textit{Affirmation and Reassurance}, \textit{Information}, \textit{Others}, \textit{Providing Suggestions}, \textit{Question}, \textit{Reflection of Feelings}, \textit{Restatement or Paraphrasing}, and \textit{Self-disclosure}. 
    % We use the publicly available \texttt{RyanDDD/empathy-strategy-classifier}, a RoBERTa-base checkpoint fine-tuned on ESConv.
    
    For each window, we compute the immediate downstream distress change $\Delta d = d(u_{t+1}) - d(u_t)$, where more negative values indicate distress reduction, and define a binary de-escalation indicator equal to 1 when $\Delta d < -0.1$. We estimate per-strategy frequency, mean $\Delta d$ with 95\% bootstrap confidence intervals, and de-escalation rate, and aggregate to the conversation level for comparison across ablation conditions (\emph{model family}, \emph{topic condition}, \emph{length condition}, \emph{label context}).
    
\paragraph{Conversation Archetype Analysis.}
    \label{sec:archetype_method}
    To test whether synthetic and real crisis conversations differ in the \emph{composition} of trajectory shapes, we cluster individual conversations into interpretable archetypes.
    % and compare archetype prevalence across sources.
    % \textit{Trajectory Representation. } 
    We summarize each interaction as a length-normalized distress trajectory by interpolating sequences onto a shared grid.
    % \textit{Pooled Clustering and Labeling. } 
    We then concatenate trajectories
    % from all three sources—zero-shot, dual-agent, and CTL, 
    and apply pooled $K$-means clustering. Centroids are assigned names based on shape features. We name each centroid from its mean distress level, total fall, and when that fall occurs, yielding five archetypes: \emph{Early Resolution}, \emph{Late Recovery}, \emph{Persistent Moderate Distress}, \emph{Steady De-escalation (high distress)}, and \emph{Unresolved High Distress}.
    % \textit{Cross Source Comparison.}
    % We contrast source-wise archetype archetype distributions, quantifying synthetic distributional shifts relative to the CTL baseline (Figure~\ref{fig:archetype}).
    % \textit{Inter-conversation Diversity.} We quantify inter-conversation diversity to test if synthetic dialogues are more repetitive or "narrow" than authentic ones. We measure the mean pairwise Euclidean distance between trajectories within each corpus, using a nonparametric bootstrap to assess significance.
    Details on the trajectory normalization and feature-based labeling of archetypes in \S\ref{app:archetype}.

\section{Results: Emotion Dynamics in CTL Grief Conversations}
\label{sec:ctl_analysis}
The following analyses use the full CTL corpus of conversations with self-identified Black or African American texters tagged as discussing grief. The annotated subset is used to validate the emotion models.
\subsection{Level 1: Utterance-Level Emotion Profiles}
\label{sec:ctl_emotion_profiles}

We focus primarily on \textbf{texter} roles, as texter emotion dynamics are the main signal of interest in crisis-support conversation analysis. Texters and volunteers nevertheless exhibit complementary emotion profiles: texters show high negative-polarity persistence (84.1\%), whereas volunteers more consistently maintain positive states, with positive-polarity persistence of 72.4\%. Volunteers' most frequent cross-emotion transition is \texttt{hopeless$\to$hopeful}, consistent with movement from acknowledging distress toward a more hopeful frame. The full role-differentiation analysis is provided in \S\ref{sec:role_diff_appendix}.

\begin{table}[t]
\centering
\small
\resizebox{\linewidth}{!}{
    \begin{tabular}{lrrlrr}
    \toprule
    \multicolumn{3}{c}{\textbf{Texter}} &
    \multicolumn{3}{c}{\textbf{Volunteer}} \\
    \cmidrule(lr){1-3}\cmidrule(lr){4-6}
    \textbf{Label} & \textbf{Count} & \textbf{\%} &
    \textbf{Label} & \textbf{Count} & \textbf{\%} \\
    \midrule
    hopeless      & 15,221 & 25.0 & hopeful       & 26,175 & 46.4 \\
    worthlessness &  8,344 & 13.7 & hopeless      & 11,866 & 21.0 \\
    hopeful       &  7,650 & 12.6 & overwhelm     &  6,253 & 11.1 \\
    overwhelm     &  5,467 &  9.0 & neutral       &  2,748 &  4.9 \\
    gratitude     &  3,942 &  6.5 & anxiety       &  1,970 &  3.5 \\
    anger         &  3,422 &  5.6 & gratitude     &  1,822 &  3.2 \\
    self          &  2,844 &  4.7 & self          &  1,424 &  2.5 \\
    anxiety       &  2,598 &  4.3 & preoccupied   &    595 &  1.1 \\
    numbness      &  2,100 &  3.5 & worry         &    590 &  1.0 \\
    sadness       &  1,722 &  2.8 & serenity      &    558 &  1.0 \\
    \bottomrule
    \end{tabular}
}
\caption{Top-10 emotion labels for CTL texter utterances
(left, $N{=}60{,}788$ labeled utterances from 2,478 conversations) and
volunteer utterances (right, $N{=}56{,}460$ labeled utterances).
For texters, 60,788 of 62,723 raw utterances received parseable labels.}
\label{tab:emotion_dist_ctl}
\end{table}

\paragraph{Emotion Distribution.} 
    Table~\ref{tab:emotion_dist_ctl} presents the top-10 emotion labels for texter and volunteer utterances in CTL dialogues.
    The texter distribution is dominated by \texttt{hopeless} (25.0\%), with the top-5 labels accounting for 66.8\% of all texter utterances. Notably, \texttt{hopeful} ranks third (12.6\%), reflecting moments of positive engagement interspersed with distress. This coexistence is particularly relevant to digital expressions of grief by Black texters, where expressions of hope and joy more broadly do not necessarily replace grief but may emerge alongside it over the course of the interaction \cite{patton-grief}.
    % \ndnote{Should distance a bit from "therapeutic", talk more about grief and cite the joy work}
    The remaining labels form a long tail, indicating that although a small set of emotions dominates the corpus, the full 37-label inventory inductively derived from the data captures a broader range of grief-related and crisis experiences. 
    % \ndnote{Discuss volunteer results} 
    Volunteer utterances show a complementary profile, with greater concentration in positive and supportive emotional states, particularly \texttt{hopeful} (46.4\%), while \texttt{hopeless} (21.0\%) and \texttt{overwhelm} (11.1\%) also remain prominent. Together, these distributions illustrate the different interactional roles occupied by texters and volunteers, while motivating the transition-based analyses below: aggregate prevalence alone does not reveal how these emotional states unfold over time.
    
    \paragraph{African American Language Analysis. }  Given that the CTL conversations involve self-identified Black texters, we note that some conversations also involve the use of African American Language (AAL)--the variety of English used by many, but not all and not exclusively, African Americans in the US \cite{grieser-black}. We qualitatively observe cases where the model appears to misinterpret the use of AAL; for example, one texter says, "\textit{I feel so alone n having to keep dis away for my kids, bout only person that bout understand I'm hurting at time is my son},"~\footnote{Note, this example is paraphrased to protect texter privacy, but the use of AAL features is conserved.} which is classified as \texttt{hopeful}. The model attributes this to the mention of the texter's son understanding, but in the original comment the texter emphasizes their loneliness and the lack of others understanding. We argue that these select cases do not significantly impact the overarching patterns identified given the validation of the model against expert annotations (\S\ref{app:model_val_appendix}) and that dense AAL features are not common in the corpus: the demographic alignment classifier introduced in \citet{blodgett-demographic} predicts AAL as the most likely label for $\sim$6\% of texts, and a probability exceeding .8 for less than 1\% of texts. We do, however, note that models' emotion labels on AAL texts are likely to be unreliable as also shown in prior work \cite{deas-aal, deas-phonate}.

    \subsection{Level 2: Micro-Dynamics}
    \label{sec:ctl_micro}
    % \subsection{Micro-Dynamics: Persistence, Pivots, and Recovery}

\paragraph{Emotion Transitions.}
Table~\ref{tab:top_transitions_ctl} presents the most frequent emotion-label transitions for CTL texter utterances. Six of the ten most frequent transitions are self-transitions, reflecting substantial emotional inertia across adjacent texter turns. Distress often persists rather than resolving immediately: \texttt{hopeless$\to$hopeless}, \texttt{worthlessness$\to$worthlessness}, and \texttt{overwhelm$\to$overwhelm} are among the most common patterns. 
        
\begin{table}[t]
\centering
\small
\begin{tabular}{lllrr}
\toprule
\multicolumn{3}{c}{\textbf{Transition}} & \textbf{Count} & \textbf{\%} \\
\midrule
hopeless      & $\to$ & hopeless      & 6,620 & 11.58 \\
worthlessness & $\to$ & worthlessness & 3,150 & 5.51 \\
hopeful       & $\to$ & hopeful       & 3,002 & 5.25 \\
worthlessness & $\to$ & hopeless      & 1,942 & 3.40 \\
overwhelm     & $\to$ & overwhelm     & 1,868 & 3.27 \\
hopeless      & $\to$ & worthlessness & 1,755 & 3.07 \\
anger         & $\to$ & anger         & 1,591 & 2.78 \\
hopeless      & $\to$ & hopeful       & 1,377 & 2.41 \\
overwhelm     & $\to$ & hopeless      & 1,299 & 2.27 \\
gratitude     & $\to$ & gratitude     & 1,257 & 2.20 \\
\bottomrule
\end{tabular}
\caption{Top-10 emotion transitions for CTL texters ($N{=}57{,}147$).
Self-transitions dominate; \texttt{hopeless$\to$hopeful} (rank~8) is the most
frequent recovery transition. Percentages are calculated over all transitions.}
\label{tab:top_transitions_ctl}
\end{table}
        
At the same time, the transition structure contains recurring movement toward hope. \texttt{Hopeless$\to$hopeful} ranks eighth overall and is the most frequent recovery transition. This pattern suggests that hope-ward movement is not limited to conversation endpoints, but also appears locally within the interaction. This also aligns with work identifying frequent discussions of self-care and joy among Black social media users discussing grief. Recovery in these CTL conversations therefore involves both persistent distress and repeated affective pivots rather than a simple replacement of negative emotion with positive emotion.

\begin{figure}[t]
    \centering
    \includegraphics[width=\linewidth]{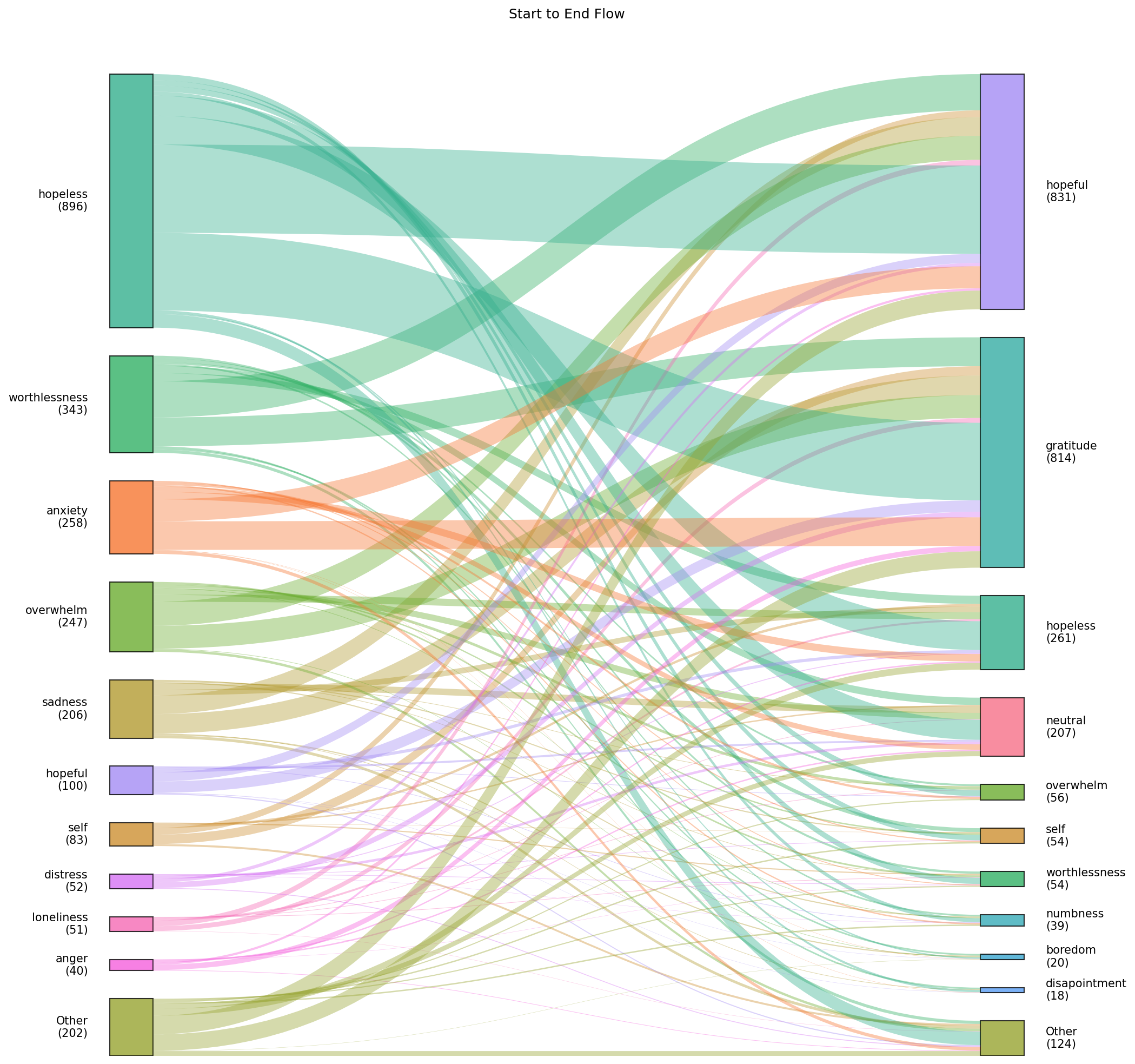}
    \caption{Conversation-start to conversation-end emotion flows for CTL texter utterances ($n=2{,}475$). Left nodes show the first \emph{substantive} emotion, after leading neutral-labelled and sub-three-word opener turns are removed; right nodes show the last. Prominent flows from \texttt{hopeless} toward \texttt{hopeful} and \texttt{gratitude} illustrate hope-ward movement at the conversation level.}
    \label{fig:ctl_sankey_emotions}
\end{figure}

Figure~\ref{fig:ctl_sankey_emotions} presents the conversation-level emotion flow from start to end, where conversation start refers to the first substantive texter emotion after leading neutral-labelled and sub-three-word opener turns are removed. Conversations beginning in \texttt{hopeless} frequently end in \texttt{hopeful} or \texttt{gratitude}, while conversations beginning in other distress-related states disperse across both positive and negative end states. This visualization complements the adjacent-turn analysis by showing the net emotional movement across entire conversations.
            
\paragraph{Emotional Volatility.} Texters also exhibit greater emotional volatility than volunteers
(Table~\ref{tab:change_rate}), consistent with fluctuating affect during crisis. Volunteers maintain more stable emotional stances across turns, further illustrating the complementary roles of the two speakers.

\paragraph{Persistence and Transition Probabilities. }
Table~\ref{tab:ctl_polarity_persistence} reports polarity persistence for CTL texters and volunteers, while Table~\ref{tab:ctl_polarity_matrix} presents the transition matrix for texter utterances.
Three key patterns emerge for CTL. First, negative states are highly persistent for texters (84.1\%), consistent with sustained distress during crisis, but substantially less so for volunteers (64.6\%).
% who actively steer away from negativity % Nick: Part of the training is focusing on validating feelings, even if negative, so I wouldn't say that they actively steer away
Second, positive states, once reached, are moderately stable for texters (64.5\%) and more so for volunteers (72.4\%), reflecting the counselor's role in anchoring conversations in a supportive frame. Third, neutral states are transient for both roles ($P(\text{stay}) \leq 25.0\%$).
For texters, neutral states transition to positive emotion in 41.4\% of cases, compared with 14.9\% of transitions from negative states, suggesting that neutral states can function as intermediate points in broader affective movement.
% \paragraph{Persistence and Transition Probabilities.}
        % Tables~\ref{tab:polarity_persistence} and~\ref{tab:polarity_matrix} present polarity persistence rates and the full row-normalized transition matrices for texter utterances in CTL and synthetic data.
        
\begin{table}[t]
\centering
\small
\begin{tabular}{lccc}
\toprule
& \multicolumn{3}{c}{$P(\text{stay})$} \\
\cmidrule(lr){2-4}
\textbf{Role} & \textbf{Neg.} & \textbf{Neut.} & \textbf{Pos.} \\
\midrule
CTL Texter    & 0.841 & 0.181 & 0.645 \\
CTL Volunteer & 0.646 & 0.250 & 0.724 \\
\bottomrule
\end{tabular}
\caption{Polarity persistence for CTL texter and volunteer utterances.}
\label{tab:ctl_polarity_persistence}
\end{table}
\begin{table}[t]
\centering
\small
\begin{tabular}{lccc}
\toprule
$\downarrow$\,\textbf{From / To}\,$\rightarrow$ &
\textbf{Neg.} & \textbf{Neut.} & \textbf{Pos.} \\
\midrule
Negative  & 0.841 & 0.010 & 0.149 \\
Neutral   & 0.405 & 0.181 & 0.414 \\
Positive  & 0.322 & 0.034 & 0.645 \\
\bottomrule
\end{tabular}
\caption{Row-normalized polarity transition matrix for CTL texter utterances.}
\label{tab:ctl_polarity_matrix}
\end{table}
\begin{table}[t]
\centering
\small
\begin{tabular}{lrr}
\toprule
\textbf{Polarity} & \textbf{Start (\%)} & \textbf{End (\%)} \\
\midrule
Negative & 90.9 & 20.5 \\
Neutral  &  0.0 &  8.6 \\
Positive &  9.1 & 70.9 \\
\bottomrule
\end{tabular}
\caption{Conversation-start and conversation-end polarity distributions for CTL texter utterances.}
\label{tab:ctl_start_end}
\end{table}

     \paragraph{Conversation Arc.}
    Table~\ref{tab:ctl_start_end} shows the starting and ending polarity distributions for CTL texter conversations, where conversation start refers to the first substantive texter utterance after leading neutral-labelled and sub-three-word opener turns are removed.
    CTL conversations overwhelmingly begin in negative states (90.9\%), while 70.9\% end in positive states.
    
    These results suggest a period of continued disclosure and emotional persistence before positive movement becomes visible in the texter's language. The CTL conversation arc is therefore not simply a difference between negative beginnings and positive endings; it is produced through repeated local transitions over the course of the interaction.

\subsection{Level 3: Macro-Dynamics}
\label{sec:ctl_macro}
    The previous analyses characterize how texter emotions evolve over the course of CTL grief conversations. We next examine the interactional role of volunteer responses: which support strategies are associated with downstream reductions in texter distress, and which are followed by continued or heightened distress?

\paragraph{Role Dynamics.}
\label{sec:role_dynamics}
    % Looking into Texter--Volunteer Interaction, 
    Table~\ref{tab:ctl_strategy_distribution} shows that CTL volunteer responses are dominated by \textit{Affirmation and Reassurance} and \textit{Question}, which together account for more than half of all predicted support strategies. This distribution reflects two central functions of crisis support: providing immediate emotional validation and eliciting enough context to understand the texter's situation. More directive or resource-oriented strategies, such as \textit{Providing Suggestions} and \textit{Information}, occur less frequently, suggesting that concrete advice and resource-sharing are used more selectively.

\begin{table}[t]
\centering
\footnotesize
\setlength{\tabcolsep}{6pt}
\begin{tabular}{lr}
\toprule
\textbf{Volunteer Strategy} & \textbf{CTL (\%)} \\
\midrule
Affirmation and Reassurance & 37.4 \\
Information                 & 3.5  \\
Others                      & 10.7 \\
Providing Suggestions       & 12.1 \\
Question                    & 24.3 \\
Reflection of Feelings      & 5.0  \\
Restatement or Paraphrasing & 6.3  \\
Self-disclosure             & 0.6  \\
\bottomrule
\end{tabular}
\caption{Distribution of predicted volunteer support strategies in CTL conversations.}
\label{tab:ctl_strategy_distribution}
\end{table}

\begin{figure}[t]
    \centering
    \includegraphics[width=\linewidth]{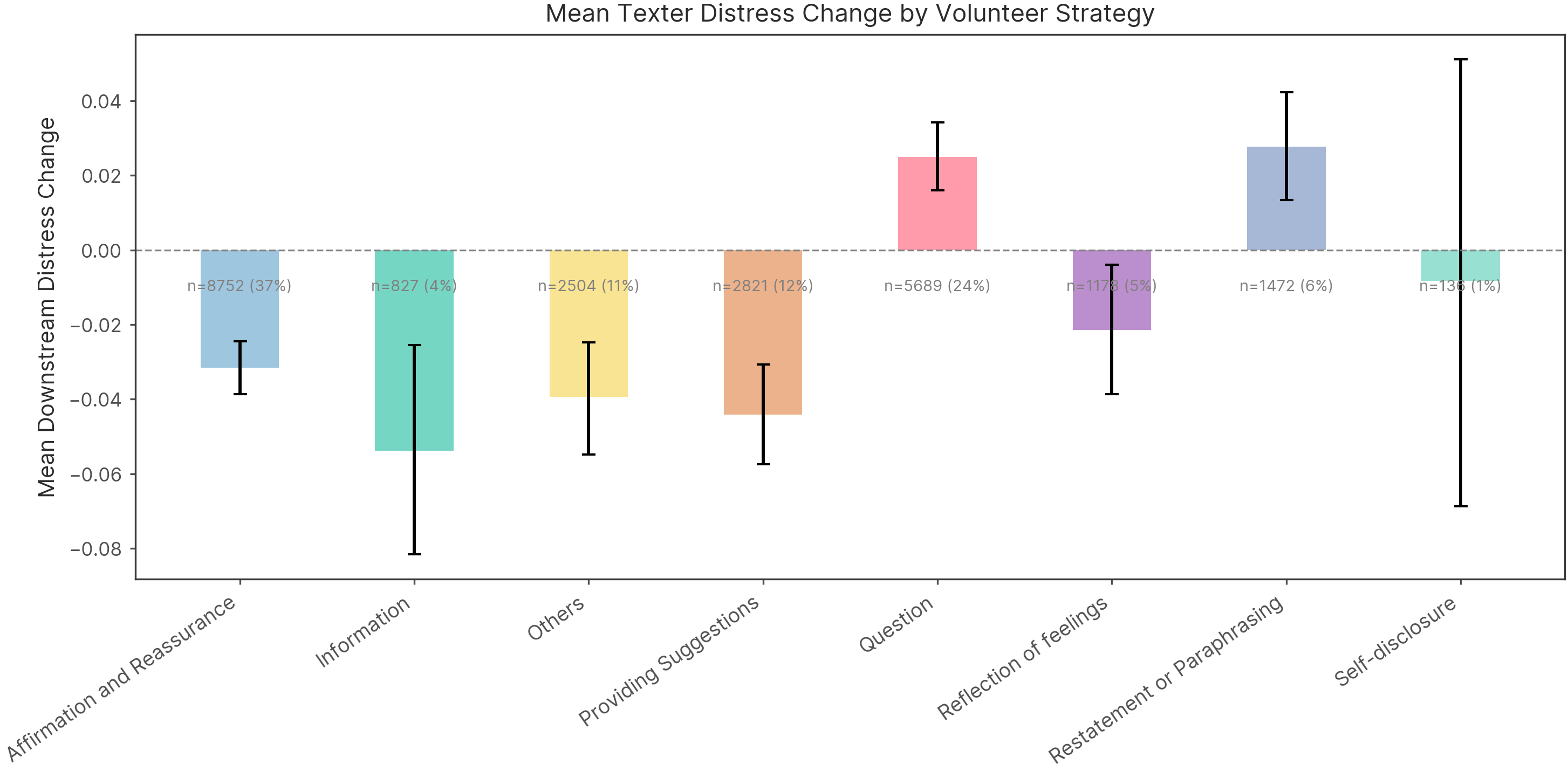}
    \caption{Volunteer strategy effects in CTL conversations. Bars show the downstream change in texter distress after each volunteer strategy, computed over local three-turn windows $(u_t, h_t, u_{t+1})$. Negative values indicate reduced texter distress after the volunteer response.}
    \label{fig:strategy_effects_ctl}
\end{figure}

We also compute the downstream change in texter distress for each strategy. Figure~\ref{fig:strategy_effects_ctl} shows that volunteer strategies differ in their association with next-turn texter distress change. Some strategies are more often followed by de-escalation, while others are associated with distress persistence or continued escalation. In particular, \textit{Information}, \textit{Providing Suggestions}, and \textit{Affirmation and Reassurance} are associated with downstream distress reduction, with \textit{Information} showing the largest mean decrease. \textit{Reflection of Feelings} is also followed by a smaller reduction in distress. However, the wider uncertainty intervals for \textit{Information} and the relatively infrequent \textit{Self-disclosure} strategy suggest that these estimates should be interpreted cautiously.

By contrast, \textit{Question} and \textit{Restatement or Paraphrasing} are followed by small mean increases in texter distress. We do not interpret this as evidence that these strategies are ineffective. Rather, in crisis-support conversations, questions and restatements often
% function to
invite elaboration, clarify the texter's situation, or keep the conversation open before de-escalation occurs. Their association with short-term distress increases may therefore reflect their role in supporting continued disclosure rather than immediately reducing distress.

These results highlight the importance of modeling crisis support as an interactional process. Volunteer strategies are not interchangeable surface forms: the same texter emotion can be followed by different support moves, and those moves are associated with different short-term affective trajectories. This role-level analysis complements the transition and archetype analyses by showing how local counselor behavior is associated with the emotional path of the conversation.

\paragraph{Conversation-Level Trajectory Archetypes.}
\label{sec:ctl_archetypes}

The trajectory analysis further shows that CTL conversations do not follow a single de-escalation pattern. Conversations are distributed relatively evenly across the identified archetypes (Fig.~\ref{fig:archetype}), with each accounting for approximately 19--22\% of the corpus. These include early resolution, late recovery, late recovery under high distress, steady de-escalation under high distress, and unresolved high distress.
This balance indicates substantial heterogeneity in how grief-related crisis conversations unfold. Some texters show early or gradual movement toward lower distress, while others recover late or remain distressed at the end of the interaction. These archetypes reinforce the micro-dynamic findings: authentic crisis support is not adequately represented by a single average recovery curve.
\paragraph{Together,} the three-level analysis shows that emotion dynamics in CTL grief conversations are neither static nor uniformly linear. Distress is highly persistent, yet conversations also contain recurring hope-ward pivots, gradual movement toward positive affect. Texters and volunteers occupy complementary emotional roles with variations in how recovery unfolds. 
These findings highlight why crisis support cannot be understood through utterance-level labels or conversation endpoints alone. \empath{} provides a unified framework for examining how affect persists, shifts, and resolves across interaction. More broadly, it offers an evaluation system for studying counseling conversations as dynamic processes rather than collections of isolated responses.

\newcommand{\commentout}[1]{}
% \section{Application to Other Domains}
% \sara{revised for arxiv}
\section{Secondary Analysis: Do Synthetic Dialogues Preserve Emotion Dynamics?}
\label{sec:synthetic_secondary}

The preceding sections use \empath{} to characterize emotion dynamics in authentic CTL grief conversations. We next use the framework to ask whether synthetic crisis-support dialogues preserve these dynamics. We generate zero-shot and dual-agent conversations using three frontier LLMs and apply the same emotion, transition, strategy, and trajectory analyses used for CTL. Full generation settings and prompts are provided in \S\ref{sec:generation_conditions}. 

% \paragraph{Similar Endpoints, Different Recovery Dynamics}
\paragraph{Micro Dynamics.}
\label{sec:synthetic_micro}

\begin{figure}[t]
    \centering
    \includegraphics[width=.95\linewidth]{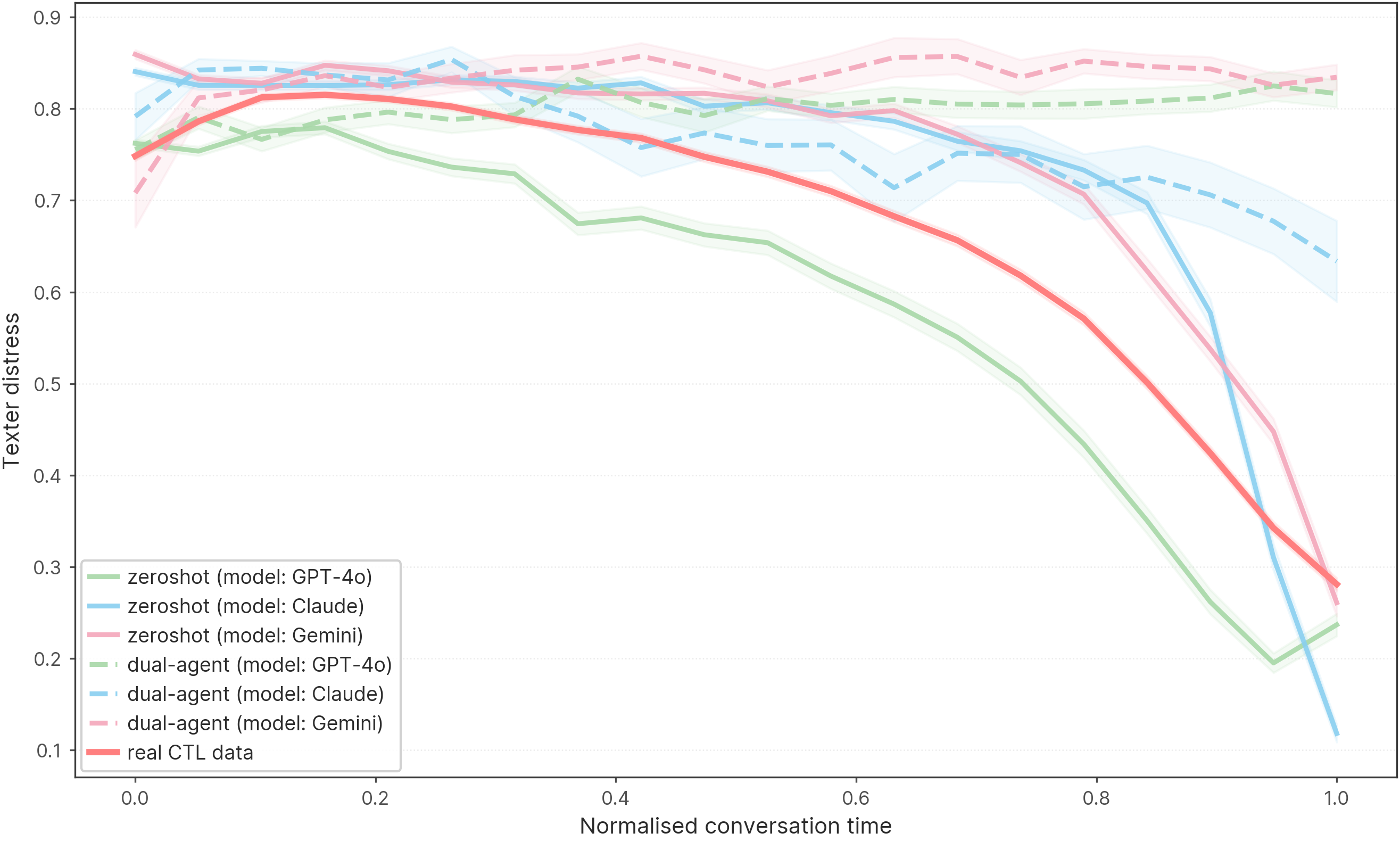}
    \caption{Mean texter distress trajectories for the full CTL analysis corpus and synthetic conditions. All conditions show distress reduction, but differ in the pacing and structure of de-escalation.}
    \label{fig:trajectory_model}
\end{figure}

CTL conversations show de-escalation composed of repeated local changes in emotional state. As Figure~\ref{fig:trajectory_model} illustrates, synthetic conversations also show substantial distress reduction, but often follow different paths: distress remains elevated for longer or changes more abruptly depending on the generation setup.

This distinction is also visible in the transition structure. CTL conversations contain recurring hope-ward pivots such as \texttt{hopeless$\to$hopeful}, alongside substantial persistence within emotional states. Synthetic texters show greater negative-state persistence than CTL texters (89.4\% vs.\ 84.1\%) and fewer direct negative$\to$positive transitions (10.3\% vs.\ 14.9\%). Their most frequent fine-grained transitions are also dominated by self-loops and movement among negative states, with no cross-polarity recovery transition appearing in the top ten. The CTL data therefore suggest that de-escalation is not simply an endpoint shift, but a sequence of smaller emotional transitions unfolding throughout the interaction. %Full transition results are provided in \S\ref{sec:synthetic_transition_appendix}.

\paragraph{Contextual Grounding.}
\label{sec:synthetic_strategy_response}

Table~\ref{tab:information_case_study} illustrates another property of the CTL interactions: support strategies are grounded in the texter's specific circumstances. Here, the CTL volunteer offers a grief-specific resource and explains why it may be relevant. The synthetic exchange uses the same broad \emph{Information} strategy but provides more generic crisis resources. This example illustrates why strategy labels alone do not capture how support is adapted to the preceding interaction.

\begin{table}[t]
\centering
\footnotesize
\renewcommand{\arraystretch}{1.3}
\begin{tabular}{p{0.95\linewidth}}
\toprule
\textbf{Paraphrased CTL Excerpt} \\
\midrule
\textbf{V:} There is also a resource made for people who have lost their child. Would you like me to share it with you? \\[2pt]
\textbf{T:} Yes, that would be helpful \\[2pt]
\textbf{V:} On \colorbox{lightbluehl}{GriefNet}, there is a community of people you can talk to that are in similar situations. \\[2pt]
\textbf{V:} I wanted to give you two options so you can figure out what resources might work best. \\[2pt]
\textbf{T:} Thank you so much! \\
\midrule
\textbf{Synthetic (zero-shot) Excerpt} \\
\midrule
\textbf{V:} At the end of this chat I can send you some resources. \\[2pt]
\textbf{T:} ok what \\[2pt]
\textbf{V:} You could text here anytime. Or call \colorbox{lightyellowhl}{988, the suicide and crisis lifeline}. Or go to an \colorbox{lightyellowhl}{ER} if it feels urgent. \\[2pt]
\textbf{T:} i dont think itll get that bad but i guess its good to know \\[2pt]
\textbf{V:} It's good to have it just in case. \\
\bottomrule
\end{tabular}
\caption{Representative \emph{Information} exchanges in CTL and synthetic conversations. The CTL volunteer offers a specific, contextually matched resource (\colorbox{lightbluehl}{GriefNet}); the synthetic volunteer offers generic escalation options (\colorbox{lightyellowhl}{988, ER}) without tailoring. Texts drawn from CTL are paraphrased.}
\label{tab:information_case_study}
\end{table}

\begin{figure}[t]
    \centering
    \includegraphics[width=\linewidth]{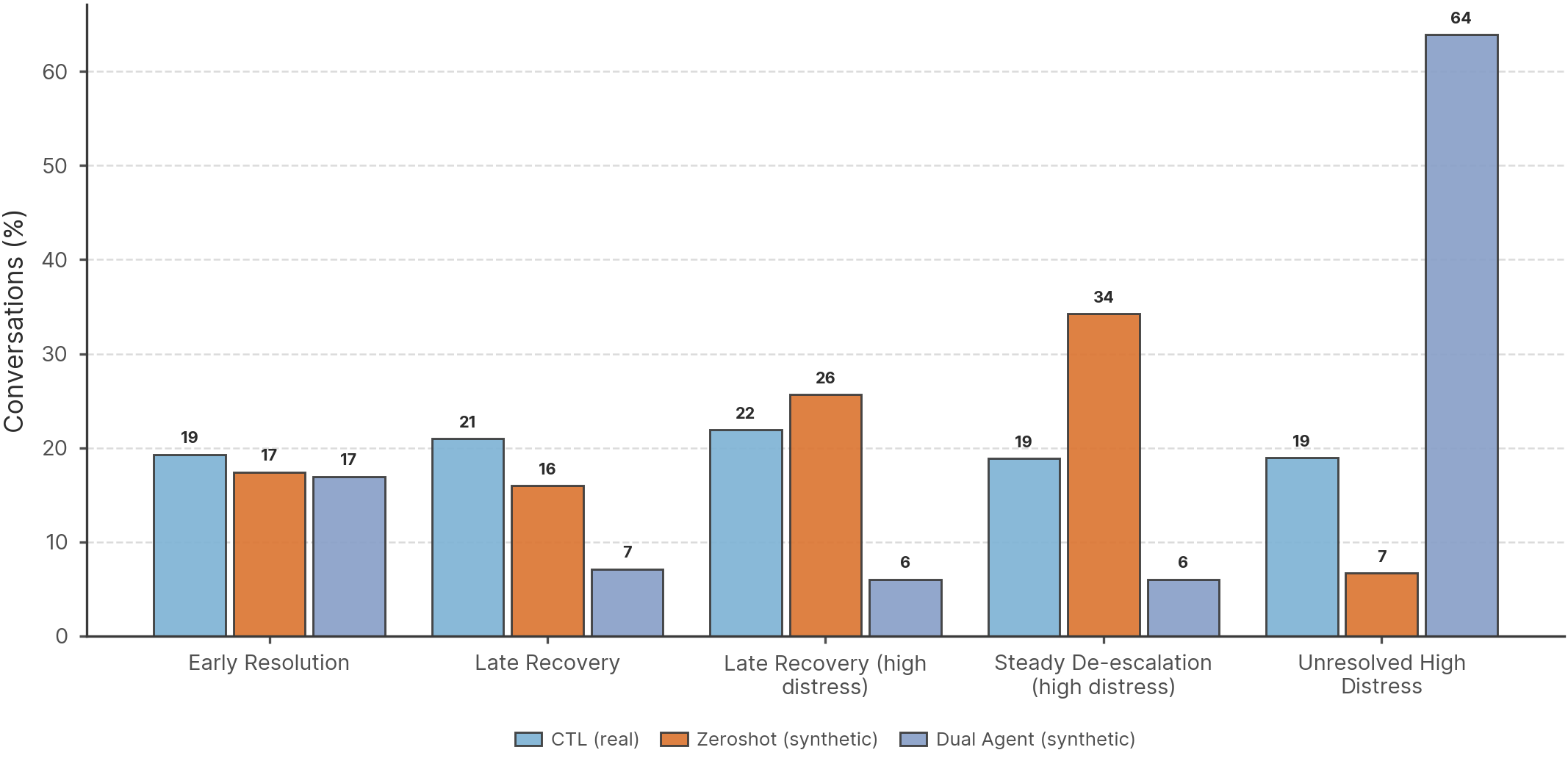}
    \caption{Distribution of conversation-level distress trajectory archetypes across the full CTL analysis corpus, zero-shot synthetic, and dual-agent synthetic conversations.}
    \label{fig:archetype}
\end{figure}

\paragraph{Macro Dynamics.}
\label{sec:synthetic_macro}

The CTL corpus also contains substantial variation at the conversation level. As Figure~\ref{fig:archetype} shows, CTL conversations are distributed relatively evenly across the five trajectory archetypes (19--22\%), reflecting heterogeneous paths through distress and recovery. Synthetic trajectories are more concentrated and depend strongly on generation setup: zero-shot conversations overrepresent \emph{Steady De-escalation (high distress)} (34\% vs.\ 19\% in CTL), whereas 64\% of dual-agent conversations fall into \emph{Unresolved High Distress}, partly due to the dual-agent termination procedure. Neither synthetic condition reproduces the balanced trajectory distribution observed in CTL. %Pairwise trajectory distances similarly indicate broader variation among CTL conversations.
% Synthetic conversations also differ from CTL at the conversation level. Figure~\ref{fig:archetype} shows that CTL conversations are distributed relatively evenly across trajectory archetypes, reflecting heterogeneous paths through distress and recovery. Zero-shot conversations overrepresent cleaner patterns such as \emph{Steady De-escalation}, while 80\% of dual-agent conversations fall into \emph{Unresolved High Distress}. Notably, the latter result is partly affected by the dual-agent termination procedure.

% Neither synthetic condition reproduces the balance of trajectory types observed in CTL. Pairwise trajectory distances provide convergent evidence: CTL conversations span a broader region of trajectory space, whereas synthetic conversations, particularly in the dual-agent condition, are substantially more concentrated. The diversity calculation is described in \S\ref{app:archetype}.

Overall, this secondary analysis further highlights the structure observed in CTL: emotional change is gradual, locally patterned, contextually grounded, and heterogeneous across conversations. Synthetic dialogues can reproduce some aggregate properties of these interactions, but do not consistently recover these finer-grained dynamics. \empath{} provides a way to characterize these properties directly rather than relying on surface plausibility alone.

\section{Conclusion}
% \sara{rewrote to end with CTL and EMPATH, not synthetic failure. defintly too long now}

We introduced \empath{}, a framework for analyzing emotion dynamics in crisis-support dialogue through utterance-level emotions, turn-level transitions, and conversation-level trajectory archetypes. Applying \empath{} to Crisis Text Line conversations with Black texters discussing grief, we find that these crisis support conversations involve persistent distress but also gradual movement toward hope, distinct texter--volunteer emotional roles, and heterogeneous recovery trajectories. These results support that computationally, crisis support is better understood as a dynamic, interactional process than as a simple shift from negative to positive emotion. Beyond CTL, we show through secondary analyses that \empath{} provides a general perspective to understand counseling dialogues, measuring how affective states persist, shift, and resolve over time.

% \begin{comment}
% Beyond CTL, we show through secondary analyses of synthetic dialogues that \empath{} provides a general perspective to understand counseling dialogues, measuring how affective states persist, shift, and resolve over time.
% \ndnote{I rephrased second paragraph to merge into previous. Also leaning toward downplaying the synthetic result in conclusion (i.e., I took the mention of the conclusion out for now).}
% % Beyond this CTL setting, \empath{} provides a general way to understand and evaluate counseling and mental health dialogues by measuring how affective states persist, shift, and resolve over time. A secondary comparison with synthetic crisis dialogues further shows that these dynamics are not fully captured by generated data, reinforcing the importance of grounding evaluation in authentic support interactions. 
% \end{comment}

While we focus primarily on CTL grief conversations in this study, in future work, \empath{} can be extended to broader counseling contexts, including peer support, therapy dialogue, counselor training simulations, and mental health dialogue systems. As privacy-preserving synthetic and simulated data become increasingly common, emotion-dynamic evaluation can help understand such dialogues and differences among them to ensure effective support. Overall, our work positions \empath{} as a tool for studying digital grief support, evaluating realism in counseling technologies, and computationally understanding broader mental health dialogues.
% \ndnote{I revised again (commented out below tho) to emphasize the focus on understanding conversations more over synthetic conversations}
% assess whether these resources preserve the temporal and interactional structure of authentic support. 
% Overall, our work positions \empath{} as a tool for studying online grief support and as a general evaluation framework for safe, realistic, and clinically meaningful counseling technologies.
% To validate synthetic, privacy-preserving alternatives for sensitive mental health data, we introduced \empath{}, a novel framework evaluating emotion dynamics across micro and macro scales of conversation. Comparing the real CTL conversations with LLM-generated dialogues reveals a pervasive "optimism bias" in synthetic data, characterized by unnaturally low emotional variability and a lack of authentic human resilience. These insights demonstrate that while current models mimic the form of counseling, they lack the affective depth required for rigorous clinical training or development of analysis tools. Future research includes bridging this gap through "failure-mode" modeling and cross-cultural validation to ensure synthetic simulations better reflect the complexity of human emotion.

\section*{Limitations}
Our analyses characterize emotional expression within text-based crisis-support conversations and do not measure longitudinal changes in grief or establish clinical recovery. Findings are specific to the samples and inclusion criteria described in Section~\ref{sec:crisis_data}; they should not be generalized to all CTL texters or to Black communities more broadly. Emotion labels are model predictions, and the single-label formulation may miss simultaneous emotions. Performance on the 493 utterances with explicit human emotion annotations does not establish equivalent performance on all utterances or on African American Language. Volunteer labels may reflect acknowledgment of texter emotions rather than volunteers’ own emotional states. Associations between support strategies and subsequent label-derived distress do not establish causal effects or counseling effectiveness. Synthetic comparisons are also sensitive to topic composition, prompt design, context-window length, and conversation termination criteria.
% Our analyses concern within-conversation emotional expression and do not establish longer-term grief recovery. Results depend on model-predicted labels, including potentially unreliable labels for AAL, and may not generalize beyond the CTL grief-tagged Black texter corpus. Associations between volunteer strategies and subsequent emotion changes do not establish causation.
% \sara{classifier uncertainty, CTL-specific setting, grief-tagged Black texter focus, no causal claims about strategies}
% \ndnote{not claiming emotion is comprehensive or that even the emotions we consider are comprehensive, focus on conversations tagged as grief}

\section*{Ethics}
Prior to data access, all Crisis Text Line data were de-identified. CTL data were accessed only under a signed Data Use Agreement and processed locally; no CTL data were provided to API-based LLMs or other online services. The example dialogue in Figure~\ref{fig:pipeline} is simulated, and CTL excerpts presented in the paper are paraphrased to protect texter privacy. No CTL data or CTL-derived material were used in synthetic generation, including the construction of prompts or topic seeds. The analyses are intended to study patterns of emotional expression and should not be treated as direct assessments of individual clinical states or of volunteer performance.
%     \ndnote{CTL data used with approved DUA, stick to local models to preserve privacy}

% \section*{Acknowledgments}

\bibliography{latex/anthology-2,latex/custom}

@ARTICLE{10.3389/fpsyt.2022.850994,
    
AUTHOR={Wilson, Da'Mere T.  and O'Connor, Mary-Frances },
           
TITLE={From Grief to Grievance: Combined Axes of Personal and Collective Grief Among Black Americans},
          
JOURNAL={Frontiers in Psychiatry},
          
VOLUME={Volume 13 - 2022},
  
YEAR={2022},
  
URL={https://www.frontiersin.org/journals/psychiatry/articles/10.3389/fpsyt.2022.850994},
  
DOI={10.3389/fpsyt.2022.850994},
  
ISSN={1664-0640}}

@article{jordan2025speech,
  title={Speech emotion recognition in mental health: Systematic review of voice-based applications},
  author={Jordan, Eric and Terrisse, Rapha{\"e}l and Lucarini, Valeria and Alrahabi, Motasem and Krebs, Marie-Odile and Descl{\'e}s, Julien and Lemey, Christophe},
  journal={JMIR mental health},
  volume={12},
  number={1},
  pages={e74260},
  year={2025},
  publisher={JMIR Publications Inc., Toronto, Canada}
}

@inproceedings{Louie_2026, series={CHI ’26},
   title={Can LLM-Simulated Practice and Feedback Upskill Human Counselors? A Randomized Study with 90+ Novice Counselors},
   url={http://dx.doi.org/10.1145/3772318.3791821},
   DOI={10.1145/3772318.3791821},
   booktitle={Proceedings of the 2026 CHI Conference on Human Factors in Computing Systems},
   publisher={ACM},
   author={Louie, Ryan and Shah, Raj Sanjay and Orney, Ifdita Hasan and Pacheco, Juan Pablo and Brunskill, Emma and Yang, Diyi},
   year={2026},
   month=Apr, pages={1–31},
   collection={CHI ’26} }

@article{Xu_2024,
   title={Mental-LLM: Leveraging Large Language Models for Mental Health Prediction via Online Text Data},
   volume={8},
   ISSN={2474-9567},
   url={http://dx.doi.org/10.1145/3643540},
   DOI={10.1145/3643540},
   number={1},
   journal={Proceedings of the ACM on Interactive, Mobile, Wearable and Ubiquitous Technologies},
   publisher={Association for Computing Machinery (ACM)},
   author={Xu, Xuhai and Yao, Bingsheng and Dong, Yuanzhe and Gabriel, Saadia and Yu, Hong and Hendler, James and Ghassemi, Marzyeh and Dey, Anind K. and Wang, Dakuo},
   year={2024},
   month=Mar, pages={1–32} }

@misc{xie2025culturalpromptingimprovesempathy,
      title={Cultural Prompting Improves the Empathy and Cultural Responsiveness of GPT-Generated Therapy Responses}, 
      author={Serena Jinchen Xie and Shumenghui Zhai and Yanjing Liang and Jingyi Li and Xuehong Fan and Trevor Cohen and Weichao Yuwen},
      year={2025},
      eprint={2512.00014},
      archivePrefix={arXiv},
      primaryClass={cs.HC},
      url={https://arxiv.org/abs/2512.00014}, 
}

@misc{louie2025care,
  title         = {Can {LLM}-Simulated Practice and Feedback Upskill Human Counselors?
                   {A} Randomized Study with 90+ Novice Counselors},
  author        = {Ryan Louie and Ifdita Hasan Orney and Juan Pablo Pacheco and
                   Raj Sanjay Shah and Emma Brunskill and Diyi Yang},
  year          = {2025},
  eprint        = {2505.02428},
  archivePrefix = {arXiv},
  primaryClass  = {cs.CL},
  url           = {https://arxiv.org/abs/2505.02428}
}

@misc{openai2024gpt4o,
  author       = {OpenAI},
  title        = {{GPT-4o} System Card},
  year         = {2024},
  url          = {https://openai.com/index/gpt-4o-system-card/}
}

@misc{anthropic2024claude,
  author       = {Anthropic},
  title        = {Claude Haiku 4.5},
  year         = {2025},
  url          = {https://www.anthropic.com/claude/haiku}
}

@misc{google2025gemini25,
  author       = {Google},
  title        = {Gemini 2.5 Pro},
  year         = {2025},
  url          = {https://deepmind.google/technologies/gemini/}
}

@inproceedings{mohammad2018obtaining,
  title     = {Obtaining Reliable Human Ratings of Valence, Arousal, and Dominance for 20,000 {English} Words},
  author    = {Mohammad, Saif M.},
  booktitle = {Proceedings of ACL},
  year      = {2018}
}

@article{llama3,
    title   = {The {L}lama 3 Herd of Models},
    author  = {Grattafiori, Aaron and Dubey, Abhimanyu and Jauhri, Abhinav and others},
    journal = {arXiv preprint arXiv:2407.21783},
    year    = {2024},
}

@article{wang2023selfconsistency,
    title   = {Self-Consistency Improves Chain of Thought Reasoning in Language Models},
    author  = {Wang, Xuezhi and Wei, Jason and Schuurmans, Dale and Le, Quoc and Chi, Ed and Narang, Sharan and Chowdhery, Aakanksha and Zhou, Denny},
    journal = {arXiv preprint arXiv:2203.11171},
    year    = {2023},
}

@inproceedings{devlin2019bert,
    title     = {{BERT}: Pre-training of Deep Bidirectional Transformers for Language Understanding},
    author    = {Devlin, Jacob and Chang, Ming-Wei and Lee, Kenton and Toutanova, Kristina},
    booktitle = {Proceedings of the 2019 Conference of the North {A}merican Chapter of the Association for Computational Linguistics: Human Language Technologies},
    year      = {2019},
    pages     = {4171--4186},
}

@article{ji2022mentalbert,
    title   = {{MentalBERT}: Publicly Available Pretrained Language Models for Mental Health},
    author  = {Ji, Shaoxiong and Zhang, Tianlin and Ansari, Luna and Fu, Jie and Tiwari, Prayag and Cambria, Erik},
    journal = {arXiv preprint arXiv:2110.15621},
    year    = {2022},
}

@article{jiang2023mistral,
    title   = {Mistral 7{B}},
    author  = {Jiang, Albert Q. and Sablayrolles, Alexandre and Mensch, Arthur and Bamford, Chris and Chaplot, Devendra Singh and Casas, Diego de las and Bressand, Florian and Lengyel, Gianna and Lample, Guillaume and Saulnier, Lucile and Lavaud, L{\'e}lio Renard and Lachaux, Marie-Anne and Stock, Pierre and Scao, Teven Le and Lavril, Thibaut and Wang, Thomas and Lacroix, Timoth{\'e}e and Sayed, William El},
    journal = {arXiv preprint arXiv:2310.06825},
    year    = {2023},
}

@misc{bai2025emotionalsupportersusemultiple,
      title={Emotional Supporters often Use Multiple Strategies in a Single Turn}, 
      author={Xin Bai and Guanyi Chen and Tingting He and Chenlian Zhou and Yu Liu},
      year={2025},
      eprint={2505.15316},
      archivePrefix={arXiv},
      primaryClass={cs.CL},
      url={https://arxiv.org/abs/2505.15316}, 
}

@inproceedings{gong2024mapping,
    title     = {A Mapping on Current Classifying Categories of Emotions Used in Multimodal Models for Emotion Recognition},
    author    = {Gong, Ziwei and Hu, Xinyi and Yao, Muyin and Zhu, Xiaoning and Hirschberg, Julia},
    booktitle = {Proceedings of the 18th Linguistic Annotation Workshop (LAW-XVIII)},
    pages     = {19--28},
    year      = {2024},
    publisher = {Association for Computational Linguistics},
}

@inproceedings{camburu-explain,
author = {Camburu, Oana-Maria and Rockt\"{a}schel, Tim and Lukasiewicz, Thomas and Blunsom, Phil},
title = {e-SNLI: natural language inference with natural language explanations},
year = {2018},
publisher = {Curran Associates Inc.},
address = {Red Hook, NY, USA},
booktitle = {Proceedings of the 32nd International Conference on Neural Information Processing Systems},
pages = {9560–9572},
numpages = {13},
location = {Montr\'{e}al, Canada},
series = {NIPS'18}
}

@inproceedings{limpijankit-counterfactual,
    title = "Counterfactual Simulatability of {LLM} Explanations for Generation Tasks",
    author = "Limpijankit, Marvin  and
      Chen, Yanda  and
      Subbiah, Melanie  and
      Deas, Nicholas  and
      McKeown, Kathleen",
    editor = "Flek, Lucie  and
      Narayan, Shashi  and
      Phương, L{\^e} Hồng  and
      Pei, Jiahuan",
    booktitle = "Proceedings of the 18th International Natural Language Generation Conference",
    month = oct,
    year = "2025",
    address = "Hanoi, Vietnam",
    publisher = "Association for Computational Linguistics",
    url = "https://aclanthology.org/2025.inlg-main.38/",
    pages = "659--683"
}

@inproceedings{flemings-privacy,
  title={Differentially Private Knowledge Distillation via Synthetic Text Generation},
  author={James Flemings and Murali Annavaram},
  booktitle={Annual Meeting of the Association for Computational Linguistics},
  year={2024},
  url={https://api.semanticscholar.org/CorpusID:268230792}
}

@article{kurakin-privacy,
  title={Harnessing large-language models to generate private synthetic text},
  author={Alexey Kurakin and Natalia Ponomareva and Umar Syed and Liam MacDermed and A. Terzis},
  journal={ArXiv},
  year={2023},
  volume={abs/2306.01684},
  url={https://api.semanticscholar.org/CorpusID:259063934}
}

@inproceedings{cabrera-synthetic,
    title = "Synthetic Empathy: Generating and Evaluating Artificial Psychotherapy Dialogues to Detect Empathy in Counseling Sessions",
    author = "Cabrera Lozoya, Daniel  and
      Hernandez Lua, Eloy  and
      Barajas Perches, Juan Alberto  and
      Conway, Mike  and
      D{'}Alfonso, Simon",
    editor = "Zirikly, Ayah  and
      Yates, Andrew  and
      Desmet, Bart  and
      Ireland, Molly  and
      Bedrick, Steven  and
      MacAvaney, Sean  and
      Bar, Kfir  and
      Ophir, Yaakov",
    booktitle = "Proceedings of the 10th Workshop on Computational Linguistics and Clinical Psychology (CLPsych 2025)",
    month = may,
    year = "2025",
    address = "Albuquerque, New Mexico",
    publisher = "Association for Computational Linguistics",
    url = "https://aclanthology.org/2025.clpsych-1.13/",
    doi = "10.18653/v1/2025.clpsych-1.13",
    pages = "157--171",
    ISBN = "979-8-89176-226-8"
}

@inproceedings{deas-masive,
    title = "{MASIVE}: Open-Ended Affective State Identification in {E}nglish and {S}panish",
    author = "Deas, Nicholas  and
      Turcan, Elsbeth  and
      Mejia, Ivan Ernesto Perez  and
      McKeown, Kathleen",
    editor = "Al-Onaizan, Yaser  and
      Bansal, Mohit  and
      Chen, Yun-Nung",
    booktitle = "Proceedings of the 2024 Conference on Empirical Methods in Natural Language Processing",
    month = nov,
    year = "2024",
    address = "Miami, Florida, USA",
    publisher = "Association for Computational Linguistics",
    url = "https://aclanthology.org/2024.emnlp-main.1139/",
    doi = "10.18653/v1/2024.emnlp-main.1139",
    pages = "20467--20485"
}

@article{hall-grief,
    author = {Christopher Hall},
    title = {Bereavement theory: recent developments in our understanding of grief and bereavement},
    journal = {Bereavement Care},
    volume = {33},
    number = {1},
    pages = {7--12},
    year = {2014},
    publisher = {Routledge},
    doi = {10.1080/02682621.2014.902610},
    URL = { 
            https://doi.org/10.1080/02682621.2014.902610
    },
    eprint = { 
            https://doi.org/10.1080/02682621.2014.902610
    }
}

@article{stroebe-dual,
    author = {Margaret Stroebe, Henk Schut},
    title = {THE DUAL PROCESS MODEL OF COPING WITH BEREAVEMENT: RATIONALE AND DESCRIPTION},
    journal = {Death Studies},
    volume = {23},
    number = {3},
    pages = {197--224},
    year = {1999},
    publisher = {Routledge},
    doi = {10.1080/074811899201046},
        note ={PMID: 10848151},
    URL = { 
            https://doi.org/10.1080/074811899201046
    },
    eprint = { 
        https://doi.org/10.1080/074811899201046
    }
}

@inbook{stroebe-meaning,
  title = {Meaning making in the dual process model of coping with bereavement.},
  ISBN = {1557987424},
  url = {http://dx.doi.org/10.1037/10397-003},
  DOI = {10.1037/10397-003},
  booktitle = {Meaning reconstruction &amp; the experience of loss.},
  publisher = {American Psychological Association},
  author = {Stroebe,  Margaret S. and Schut,  Henk},
  year = {2001},
  pages = {55–73}
}

@article{neimeyer-meaning,
  title = {Mourning and Meaning},
  volume = {46},
  ISSN = {1552-3381},
  url = {http://dx.doi.org/10.1177/000276402236676},
  DOI = {10.1177/000276402236676},
  number = {2},
  journal = {American Behavioral Scientist},
  publisher = {SAGE Publications},
  author = {Neimeyer,  Robert A. and Prigerson,  Holly G. and Davies,  BETTY},
  year = {2002},
  month = Oct,
  pages = {235–251}
}

@article{moore-digital,
  title = {Social Media Mourning: Using Grounded Theory to Explore How People Grieve on Social Networking Sites},
  volume = {79},
  ISSN = {1541-3764},
  url = {http://dx.doi.org/10.1177/0030222817709691},
  DOI = {10.1177/0030222817709691},
  number = {3},
  journal = {OMEGA - Journal of Death and Dying},
  publisher = {SAGE Publications},
  author = {Moore,  Jensen and Magee,  Sara and Gamreklidze,  Ellada and Kowalewski,  Jennifer},
  year = {2017},
  month = May,
  pages = {231–259}
}

@InProceedings{patton-grief,
author="Patton, Desmond U.
and Kleiner, Shana
and Miller, Shug
and Deas, Nicholas
and Edwards, Fahnmusa J.
and Grieser, Jessica A.
and Shepard, James
and Turcan, Elsbeth
and McKeown, Kathleen",
editor="de la Iglesia, Daniel H.
and de Paz Santana, Juan F.
and L{\'o}pez Rivero, Alfonso J.",
title="Digital Narratives of Grief and Resilience: Insights from the Integrating Emotional Stories Online (IESO) Platform",
booktitle="New Trends in Disruptive Technologies, Tech Ethics and Artificial Intelligence",
year="2025",
publisher="Springer Nature Switzerland",
address="Cham",
pages="368--378",
isbn="978-3-031-99474-6"
}

@inproceedings{deas-aal,
    title = "Evaluation of {A}frican {A}merican Language Bias in Natural Language Generation",
    author = "Deas, Nicholas  and
      Grieser, Jessica  and
      Kleiner, Shana  and
      Patton, Desmond  and
      Turcan, Elsbeth  and
      McKeown, Kathleen",
    editor = "Bouamor, Houda  and
      Pino, Juan  and
      Bali, Kalika",
    booktitle = "Proceedings of the 2023 Conference on Empirical Methods in Natural Language Processing",
    month = dec,
    year = "2023",
    address = "Singapore",
    publisher = "Association for Computational Linguistics",
    url = "https://aclanthology.org/2023.emnlp-main.421/",
    doi = "10.18653/v1/2023.emnlp-main.421",
    pages = "6805--6824"
}

@inproceedings{
    deas-phonate,
    title={Phon{AT}e: Impact of Type-Written Phonological Features of African American Language on Generative Language Modeling Tasks},
    author={Nicholas Deas and Jessica A Grieser and Xinmeng Hou and Shana Kleiner and Tajh Martin and Sreya Nandanampati and Desmond U. Patton and Kathleen McKeown},
    booktitle={First Conference on Language Modeling},
    year={2024},
    url={https://openreview.net/forum?id=rXEwxmnGQs}
}

@book{grieser-black,
  title={The Black side of the river: Race, language, and belonging in Washington, DC},
  author={Grieser, Jessica A},
  year={2022},
  publisher={Georgetown University Press}
}

@inproceedings{blodgett-demographic,
    title = "Demographic Dialectal Variation in Social Media: A Case Study of {A}frican-{A}merican {E}nglish",
    author = "Blodgett, Su Lin  and
      Green, Lisa  and
      O{'}Connor, Brendan",
    editor = "Su, Jian  and
      Duh, Kevin  and
      Carreras, Xavier",
    booktitle = "Proceedings of the 2016 Conference on Empirical Methods in Natural Language Processing",
    month = nov,
    year = "2016",
    address = "Austin, Texas",
    publisher = "Association for Computational Linguistics",
    url = "https://aclanthology.org/D16-1120/",
    doi = "10.18653/v1/D16-1120",
    pages = "1119--1130"
}

@inproceedings{liu-etal-2026-review,
    title = "A Review of Incorporating Psychological Theories in {LLM}s",
    author = "Liu, Zizhou  and
      Gong, Ziwei  and
      Ai, Lin  and
      Hui, Zheng  and
      Chen, Run  and
      Leach, Colin Wayne  and
      Greene, Michelle R.  and
      Hirschberg, Julia",
    editor = "Demberg, Vera  and
      Inui, Kentaro  and
      Marquez, Llu{\'i}s",
    booktitle = "Proceedings of the 19th Conference of the {E}uropean Chapter of the {A}ssociation for {C}omputational {L}inguistics (Volume 1: Long Papers)",
    month = mar,
    year = "2026",
    address = "Rabat, Morocco",
    publisher = "Association for Computational Linguistics",
    url = "https://aclanthology.org/2026.eacl-long.350/",
    doi = "10.18653/v1/2026.eacl-long.350",
    pages = "7459--7495",
    ISBN = "979-8-89176-380-7"
}

@inproceedings{zhang-etal-2025-intentionesc,
    title = "{I}ntention{ESC}: An Intention-Centered Framework for Enhancing Emotional Support in Dialogue Systems",
    author = "Zhang, Xinjie  and
      Wang, Wenxuan  and
      Jin, Qin",
    editor = "Che, Wanxiang  and
      Nabende, Joyce  and
      Shutova, Ekaterina  and
      Pilehvar, Mohammad Taher",
    booktitle = "Findings of the Association for Computational Linguistics: ACL 2025",
    month = jul,
    year = "2025",
    address = "Vienna, Austria",
    publisher = "Association for Computational Linguistics",
    url = "https://aclanthology.org/2025.findings-acl.1358/",
    doi = "10.18653/v1/2025.findings-acl.1358",
    pages = "26494--26516",
    ISBN = "979-8-89176-256-5"
}

@inproceedings{wan-etal-2025-emodynamix,
    title = "{E}mo{D}ynami{X}: Emotional Support Dialogue Strategy Prediction by Modelling {M}i{X}ed Emotions and Discourse Dynamics",
    author = "Wan, Chenwei  and
      Labeau, Matthieu  and
      Clavel, Chlo{\'e}",
    editor = "Chiruzzo, Luis  and
      Ritter, Alan  and
      Wang, Lu",
    booktitle = "Proceedings of the 2025 Conference of the Nations of the Americas Chapter of the Association for Computational Linguistics: Human Language Technologies (Volume 1: Long Papers)",
    month = apr,
    year = "2025",
    address = "Albuquerque, New Mexico",
    publisher = "Association for Computational Linguistics",
    url = "https://aclanthology.org/2025.naacl-long.81/",
    doi = "10.18653/v1/2025.naacl-long.81",
    pages = "1678--1695",
    ISBN = "979-8-89176-189-6"
}

@inproceedings{liu-etal-2021-towards,
    title = "Towards Emotional Support Dialog Systems",
    author = "Liu, Siyang  and
      Zheng, Chujie  and
      Demasi, Orianna  and
      Sabour, Sahand  and
      Li, Yu  and
      Yu, Zhou  and
      Jiang, Yong  and
      Huang, Minlie",
    editor = "Zong, Chengqing  and
      Xia, Fei  and
      Li, Wenjie  and
      Navigli, Roberto",
    booktitle = "Proceedings of the 59th Annual Meeting of the Association for Computational Linguistics and the 11th International Joint Conference on Natural Language Processing (Volume 1: Long Papers)",
    month = aug,
    year = "2021",
    address = "Online",
    publisher = "Association for Computational Linguistics",
    url = "https://aclanthology.org/2021.acl-long.269/",
    doi = "10.18653/v1/2021.acl-long.269",
    pages = "3469--3483"
}

@inproceedings{badawi-etal-2026-trust,
    title = "When Can We Trust {LLM}s in Mental Health? Large-Scale Benchmarks for Reliable {LLM} Evaluation",
    author = "Badawi, Abeer  and
      Rahimi, Elahe  and
      Laskar, Md Tahmid Rahman  and
      Grach, Sheri  and
      Bertrand, Lindsay  and
      Danok, Lames  and
      Dhanesh, Prathiba  and
      Huang, Jimmy  and
      Rudzicz, Frank  and
      Dolatabadi, Elham",
    editor = "Demberg, Vera  and
      Inui, Kentaro  and
      Marquez, Llu{\'i}s",
    booktitle = "Proceedings of the 19th Conference of the {E}uropean Chapter of the {A}ssociation for {C}omputational {L}inguistics (Volume 1: Long Papers)",
    month = mar,
    year = "2026",
    address = "Rabat, Morocco",
    publisher = "Association for Computational Linguistics",
    url = "https://aclanthology.org/2026.eacl-long.180/",
    doi = "10.18653/v1/2026.eacl-long.180",
    pages = "3873--3896",
    ISBN = "979-8-89176-380-7"
}

@inproceedings{wu-etal-2025-multimodal,
    title = "Multimodal Emotion Recognition in Conversations: A Survey of Methods, Trends, Challenges and Prospects",
    author = "Wu, ChengYan  and
      Cai, Yiqiang  and
      Liu, Yang  and
      Zhu, Pengxu  and
      Xue, Yun  and
      Gong, Ziwei  and
      Hirschberg, Julia  and
      Ma, Bolei",
    editor = "Christodoulopoulos, Christos  and
      Chakraborty, Tanmoy  and
      Rose, Carolyn  and
      Peng, Violet",
    booktitle = "Findings of the Association for Computational Linguistics: EMNLP 2025",
    month = nov,
    year = "2025",
    address = "Suzhou, China",
    publisher = "Association for Computational Linguistics",
    url = "https://aclanthology.org/2025.findings-emnlp.332/",
    doi = "10.18653/v1/2025.findings-emnlp.332",
    pages = "6257--6274",
    ISBN = "979-8-89176-335-7"
}

@inproceedings{tripodi-etal-2025-assessing,
    title = "Assessing effective de-escalation of crisis conversations using transformer-based models and trend statistics",
    author = "Tripodi, Ignacio J.  and
      Buda, Greg  and
      Meagher, Margaret  and
      Olson, Elizabeth A.",
    editor = "Christodoulopoulos, Christos  and
      Chakraborty, Tanmoy  and
      Rose, Carolyn  and
      Peng, Violet",
    booktitle = "Proceedings of the 2025 Conference on Empirical Methods in Natural Language Processing",
    month = nov,
    year = "2025",
    address = "Suzhou, China",
    publisher = "Association for Computational Linguistics",
    url = "https://aclanthology.org/2025.emnlp-main.1512/",
    doi = "10.18653/v1/2025.emnlp-main.1512",
    pages = "29763--29777",
    ISBN = "979-8-89176-332-6"
}

@inproceedings{louie-etal-2024-roleplay,
    title = "Roleplay-doh: Enabling Domain-Experts to Create {LLM}-simulated Patients via Eliciting and Adhering to Principles",
    author = "Louie, Ryan  and
      Nandi, Ananjan  and
      Fang, William  and
      Chang, Cheng  and
      Brunskill, Emma  and
      Yang, Diyi",
    editor = "Al-Onaizan, Yaser  and
      Bansal, Mohit  and
      Chen, Yun-Nung",
    booktitle = "Proceedings of the 2024 Conference on Empirical Methods in Natural Language Processing",
    month = nov,
    year = "2024",
    address = "Miami, Florida, USA",
    publisher = "Association for Computational Linguistics",
    url = "https://aclanthology.org/2024.emnlp-main.591/",
    doi = "10.18653/v1/2024.emnlp-main.591",
    pages = "10570--10603"
}

@inproceedings{buda-etal-2026-extraction,
    title = "Extraction of Texters' Explicit Emotion Expressions in Crisis Conversations",
    author = "Buda, Greg  and
      Tripodi, Ignacio J.  and
      Zuromski, Kelly L.  and
      Meagher, Margaret  and
      Olson, Elizabeth A.",
    editor = "Liakata, Maria  and
      Moreira, Viviane P.  and
      Zhang, Jiajun  and
      Jurgens, David",
    booktitle = "Findings of the {A}ssociation for {C}omputational {L}inguistics: {ACL} 2026",
    month = jul,
    year = "2026",
    address = "San Diego, California, United States",
    publisher = "Association for Computational Linguistics",
    url = "https://aclanthology.org/2026.findings-acl.2/",
    doi = "10.18653/v1/2026.findings-acl.2",
    pages = "27--44",
    ISBN = "979-8-89176-395-1"
}

@inproceedings{gong-etal-2023-eliciting,
    title = "Eliciting Rich Positive Emotions in Dialogue Generation",
    author = "Gong, Ziwei  and
      Min, Qingkai  and
      Zhang, Yue",
    editor = "Chawla, Kushal  and
      Shi, Weiyan",
    booktitle = "Proceedings of the First Workshop on Social Influence in Conversations (SICon 2023)",
    month = jul,
    year = "2023",
    address = "Toronto, Canada",
    publisher = "Association for Computational Linguistics",
    url = "https://aclanthology.org/2023.sicon-1.1/",
    doi = "10.18653/v1/2023.sicon-1.1",
    pages = "1--8"
}

@inproceedings{wu-etal-2024-multimodal,
    title = "Multimodal Multi-loss Fusion Network for Sentiment Analysis",
    author = "Wu, Zehui  and
      Gong, Ziwei  and
      Koo, Jaywon  and
      Hirschberg, Julia",
    editor = "Duh, Kevin  and
      Gomez, Helena  and
      Bethard, Steven",
    booktitle = "Proceedings of the 2024 Conference of the North American Chapter of the Association for Computational Linguistics: Human Language Technologies (Volume 1: Long Papers)",
    month = jun,
    year = "2024",
    address = "Mexico City, Mexico",
    publisher = "Association for Computational Linguistics",
    url = "https://aclanthology.org/2024.naacl-long.197/",
    doi = "10.18653/v1/2024.naacl-long.197",
    pages = "3588--3602"
}

@inproceedings{wang-etal-2024-patient,
    title = "{PATIENT}-$\psi$: Using Large Language Models to Simulate Patients for Training Mental Health Professionals",
    author = "Wang, Ruiyi  and
      Milani, Stephanie  and
      Chiu, Jamie C.  and
      Zhi, Jiayin  and
      Eack, Shaun M.  and
      Labrum, Travis  and
      Murphy, Samuel M  and
      Jones, Nev  and
      Hardy, Kate V  and
      Shen, Hong  and
      Fang, Fei  and
      Chen, Zhiyu",
    editor = "Al-Onaizan, Yaser  and
      Bansal, Mohit  and
      Chen, Yun-Nung",
    booktitle = "Proceedings of the 2024 Conference on Empirical Methods in Natural Language Processing",
    month = nov,
    year = "2024",
    address = "Miami, Florida, USA",
    publisher = "Association for Computational Linguistics",
    url = "https://aclanthology.org/2024.emnlp-main.711/",
    doi = "10.18653/v1/2024.emnlp-main.711",
    pages = "12772--12797"
}

\clearpage
% \onecolumn
\appendix

% \section{Appendix}

% \section{Example Appendix}
% \label{sec:appendix}

\section{Emotion Detection Prompt, Label Inventory, and Experimental Set-Ups}
\label{sec:prompt_appendix}
\paragraph{Data Pre-processing.}
% \msnote{This paragraph could go in the appendix if you need to save space} \wen{Keeping in main text for now since it explains a design decision that affects results. Can move to appendix if space is tight.} 
Before inference, we exclude three categories of utterances: (i) system messages, (ii) single-character Y/N texter feedback, and (iii) trivial first texter messages, defined as the first texter utterance in a conversation when it is empty, whitespace-only, or consists of a single token (e.g., ``Hi''). These trivial openers carry little or no emotional content and would otherwise receive unreliable \emph{neutral} labels, potentially distorting downstream transition statistics.

For sequence-level analyses, we apply an additional opener filter after inference. Starting from the beginning of each conversation, leading utterances are removed iteratively until reaching the first utterance that is both non-\emph{neutral} and substantive. An utterance is considered non-substantive if it contains fewer than three whitespace-delimited words or, for space-free CJK text, at most three characters. This criterion is necessary because short procedural openers such as ``HOME'', ``WARM'', or ``Ok'' may receive non-neutral labels due primarily to surrounding conversational context rather than their own semantic content. Importantly, this filtering is restricted to conversation openers: mid-conversation utterances are never removed, and all mid-conversation \emph{neutral} labels are retained. 

This second filter is applied only to sequence-level statistics. Label-frequency analyses are computed over all labeled utterances after the initial pre-processing step, since removing conversation openers would alter the overall label distribution and therefore bias what is intended to be a corpus-level register statistic.

% \sararevision{Appendix~\ref{sec:prompt_appendix}: verify the filtered-message count and reconcile the inference total below with Table~\ref{tab:emotion_dist_ctl} and Appendix~\ref{sec:context_comparison_appendix}; distinguish preprocessing exclusions from unparseable labels.}

\paragraph{Experimental Setups. }
\label{app:setups}
We use meta-llama/Llama-3.2-3B-Instruct \cite{llama3} as the backbone model, loaded in FP16 precision on a single CUDA-capable GPU. Inference uses greedy decoding (Stage~1) with a maximum of 128 new tokens. For self-consistency runs (Stage~3), we sample with temperature~$=0.7$ and top-$p{=}0.9$ over $K{=}5$ independent generations. The batch size is set to 64 and the random seed to 42 for reproducibility. Total inference time for the CTL dataset was approximately 27.5 hours (117,248 items across 1,832 batches).

We run two configurations per corpus: (i)~\emph{without context} (single-utterance input), and (ii)~\emph{with context} using a sliding window of the 5~most recent preceding utterances. In both settings, system messages, Y/N feedback rows, and trivial first texter messages are excluded from inference. For the synthetic data, we use context length 10 to match the longer average turn lengths in generated conversations.

\paragraph{Prompt Template (with-context mode). }
The following template is instantiated per utterance. \texttt{\{label\_definitions\}} is expanded to the full 37-label list with definitions; \texttt{\{conversation\}} contains the sliding window of preceding utterances plus the target; \texttt{\{author\}} is the speaker role of the target utterance; \texttt{\{label\_list\}} is the comma-separated list of allowed labels.

\begin{small}
\begin{verbatim}
You are an expert in emotion recognition
and mental health support.
Below is a conversation between a texter
and a volunteer. Here are the possible
emotion labels:

{label_definitions}

Conversation so far:
{conversation}

Focus especially on the last message by the
{author} shown above.
When determining the most appropriate emotion
label, give the most weight to the content
and tone of the current (last) message,
and only use the earlier conversation as
supporting context if needed.

You MUST respond in exactly this format
and nothing else:

Label: <one label chosen from this list:
        {label_list}>
Reason: <one short sentence, referencing
         the definition and the current
         message>
\end{verbatim}
\end{small}

\paragraph{Label Inventory. }
Table~\ref{tab:emotion_labels} lists all 37 emotion labels and their definitions as provided to the model.

\begin{table*}[t]
\centering
\small
\begin{tabular}{ll}
\hline
\textbf{Label} & \textbf{Definition (abridged)} \\
\hline
neutral & No strong positive or negative emotion; matter-of-fact tone \\
sadness & Feeling sad, down \\
fear & Feeling afraid, scared \\
anxiety & Feeling anxious or symptoms of generalised anxiety \\
joy & Long-lasting contentment and satisfaction with life \\
love & Feeling loved, or loving others \\
happiness & Fleeting emotion sparked by a particular moment \\
hopeful & Feeling a sense of hope / looking toward the future \\
hopeless & Feeling a lack of hope; not positive about future plans \\
guilt & Feeling guilty \\
loneliness & Feeling lonely, isolated, away from others \\
shame & Feeling shameful \\
anger & Feeling angry, furious \\
longing & Feeling longing toward someone or something \\
numbness & Feeling numb, could not feel anything \\
disapointment & Feeling disappointed or frustrated \\
stress & Feeling stress due to life events \\
worthlessness & Feeling worthless, like a failure \\
self & Feeling self-doubt or being self-aware \\
gratitude & Feeling grateful for life, community, things \\
disgust & Feeling disgusted \\
anticipation & Anticipating \\
resilient & Feeling resilient \\
distraction & Feeling distracted \\
surprise & Feeling surprised, out of expectation \\
boredom & Feeling bored \\
trust & Feeling trusted or trustworthy toward others \\
tired & Feeling tired and exhausted \\
overwhelm & Feeling intense emotions or overwhelmed \\
chaotic & Feeling chaotic, things out of control \\
mood & Having mood swings \\
preoccupied & Feeling preoccupied or busy \\
distress & General feelings of upset / emotional dysregulation \\
worry & Feeling worried or concerned \\
empathy & Feeling empathetic toward others \\
regret & Feeling regret \\
serenity & Feeling serene, calm, peaceful \\
\hline
\end{tabular}
\caption{The 37 emotion labels and abridged definitions used in the detection prompt. Full definitions are provided verbatim to the model. The \texttt{self} label merges the original coding-book categories \texttt{self-doubt} and \texttt{self-aware}.}
\label{tab:emotion_labels}
\end{table*}

\ndnote{Think we can move this (the cascading strategy description) to Appendix or at least just summarize here.} \ndnote{Moved to appendix for now}
\paragraph{Emotion Detection Pipeline.}
\label{sec:emotion_detection_pipeline}
We use a three-stage cascading procedure, applied independently to each batch so that later stages process only unresolved or uncertain examples:
\begin{enumerate}
    \setlength{\itemsep}{0pt}
    \setlength{\parskip}{0pt}
    \setlength{\parsep}{0pt}

    \item \textbf{Deterministic generation.}
    The model generates greedily (temperature~$=0$). A prediction is accepted only from an explicitly completed \texttt{Label:} field. The generated value is matched to one of the 37 valid labels after normalization; if no exact match is found, we accept a valid label appearing as the leading whole word (e.g., ``anxiety, because\ldots''). Unfilled placeholders and outputs listing three or more distinct labels are rejected. We do not perform whole-output substring matching.

    \item \textbf{First-token scoring.}
    For rows that fail parsing, we score all labels in a single forward pass using the logits at the \texttt{Label:} position and select the label with the highest first-token score.

    \item \textbf{Self-consistency refinement.}
    If the margin between the top two scores is small ($\Delta < 1.0$), we draw $K{=}5$ stochastic generations (temperature~$=0.7$, top-$p{=}0.9$) and use majority vote \cite{wang2023selfconsistency}. Unparseable samples are discarded; if none parse, the stage-2 prediction is retained. Uncertain rows are batched together for each of the $K$ sampling passes.
\end{enumerate}

If a row reaches a fallback stage when no local model is available, it is assigned \emph{neutral} and explicitly flagged as a fallback case.
% \clearpage

% \clearpage
\section{Model Validation}
\label{app:model_val_appendix}
We validate the emotion detection model by comparing three candidate architectures on the 100-conversation CTL annotation sample. Each model's predictions are evaluated against human reference annotations from six independently coded CTL conversation subsets using two complementary metrics: \emph{semantic similarity} (cosine similarity between label embeddings) and \emph{exact-match accuracy}. We favor semantic similarity as the primary metric because, with a 37-category inventory, exact match is overly strict: clinically similar predictions (e.g., \texttt{hopeless} vs.\ \texttt{worthlessness}) are penalized equally to entirely wrong ones (e.g., \texttt{hopeless} vs.\ \texttt{joy}). Semantic similarity provides graded credit that better reflects the practical quality of predictions. Standard per-class precision, recall, and F1 are not reported because (i)~BERT Emotions uses a different label taxonomy, making class-level comparison infeasible, and (ii)~with 37 fine-grained categories and limited per-subset sizes (503--2,203 utterances), many individual classes have too few samples for stable per-class estimates. All models are evaluated on utterances with explicit human emotion annotations.
% \sararevision{Appendix~\ref{app:model_val_appendix}: reconcile the subset-size range stated here with the explicitly labeled evaluation count in Table~\ref{tab:multi_level} and the annotation counts in Table~\ref{tab:data-stats}; clarify which utterances enter validation.}

\paragraph{Candidate Models. }
We compare five approaches spanning the encoder-only and generative paradigms:
\begin{enumerate}
    \item \textbf{BERT Emotions}~\cite{devlin2019bert}: A BERT-base model fine-tuned on emotion classification, serving as the reference baseline. Because BERT Emotions uses a different label taxonomy than our 37-category inventory, exact-match accuracy cannot be computed; only semantic similarity between its predicted labels and the reference is reported.
    \item \textbf{MentalBERT}~\cite{ji2022mentalbert}: A BERT variant pre-trained on mental-health corpora (Reddit counseling, psychological forums), hypothesized to better capture therapeutic language.
    \item \textbf{Llama-3.2-3B-Instruct}: The pipeline described in \S\ref{sec:emotion_detection_pipeline}, using Llama-3.2-3B-Instruct with retrieval-augmented label definitions. Evaluated under two context conditions: without conversational context (single-utterance input) and with context (sliding window of preceding turns).
    \item \textbf{Llama-3.1-8B-Instruct}: A larger Llama variant (8B parameters) to assess whether increased model capacity improves emotion classification within the same pipeline.
    \item \textbf{Mistral-7B-Instruct-v0.3}~\cite{jiang2023mistral}: A 7B-parameter instruction-tuned model from a different model family, included to evaluate cross-architecture generalization of the pipeline.
\end{enumerate}

\paragraph{Comparison Results. }
BERT Emotions and MentalBERT are both BERT-based architectures with a maximum input length of 512 tokens (64 tokens for BERT Emotions), designed for single-sentence classification. They cannot naturally incorporate multi-turn conversational context: BERT Emotions was fine-tuned on short single-sentence inputs, and MentalBERT was continually pre-trained on individual Reddit posts rather than multi-turn dialogues. In contrast, the generative models (Llama-3.2-3B, Llama-3.1-8B, Mistral-7B) support large context windows and were instruction-tuned on conversational formats. We compare all five models in the without-context setting for a fair evaluation. Table~\ref{tab:multi_level} reports performance on 493 human-annotated utterances with explicit emotion labels.

Mistral-7B-Instruct-v0.3 achieves the highest accuracy at all taxonomy levels. However, we use Llama-3.2-3B-Instruct for all primary analyses. The accuracy gap narrows as the taxonomy becomes coarser: at the Ekman 7-category level the difference is only 1.5 percentage points (0.682 vs.\ 0.667), and at the sentiment level the two models are nearly indistinguishable (0.844 vs.\ 0.842). Moreover, Llama-3.2-3B-Instruct achieves semantic similarity comparable to or higher than the other generative models at every mapped level (0.975--0.983), indicating that its predictions are semantically closest to the human annotations even when the exact label differs. The accuracy differences therefore reflect fine-grained synonym disagreements (e.g., \texttt{hopeless} vs.\ \texttt{worthlessness}) rather than systematic misclassification. Because our dynamics analysis aggregates over tens of thousands of transitions, these per-utterance differences are unlikely to alter corpus-level patterns such as transition rankings, polarity persistence rates, or conversation-arc statistics. Additionally, Llama-3.2-3B (3B parameters) outperforms the larger Llama-3.1-8B (8B parameters) at all taxonomy levels despite being less than half the size, offering the best accuracy-to-compute tradeoff for processing 117K+ utterances across both CTL and synthetic corpora. The multi-model validation thus serves to demonstrate that the pipeline generalizes across model families and scales, rather than to select a single backbone.
% \sararevision{Appendix~\ref{app:model_val_appendix}: verify the semantic-similarity range and model comparisons against Table~\ref{tab:multi_level}; update the score descriptions and bolded best-score markers wherever they conflict.}

\paragraph{Multi-Level Taxonomy Evaluation. }
The 37-label exact-match accuracy reported above is a conservative lower bound: clinically similar predictions (e.g., \texttt{hopeless} vs.\ \texttt{worthlessness}) are penalized as errors even though both reflect the same broad emotional state. To quantify how much of the apparent ``error'' is attributable to fine-grained label confusion rather than genuine misclassification, we evaluate accuracy at progressively coarser emotion taxonomies.

We adopt the cascading mapping method of \citet{gong2024mapping}, which maps fine-grained emotion labels to coarser categories based on shared names, valence/arousal similarity, and human evaluation. Our 37 labels are first mapped to Plutchik's 14 fine-grained categories (e.g., \texttt{hopeless}, \texttt{worthlessness}, \texttt{loneliness} $\to$ \emph{sadness}; \texttt{anxiety}, \texttt{stress}, \texttt{overwhelm} $\to$ \emph{fear}; \texttt{happiness}, \texttt{love}, \texttt{gratitude} $\to$ \emph{joy}), then cascaded to Ekman's 7 basic emotions and 3 sentiment classes following the paper's validated hierarchy. The full mapping is provided in \S\ref{sec:multi_level_mapping_appendix}. Both the model's predicted labels and the human reference annotations are mapped to the same coarser taxonomy before computing accuracy, so a prediction of \texttt{hopeless} against a human label of \texttt{worthlessness} counts as correct at the 14-category level (both map to \emph{sadness}) even though it is an error at the 37-label level.

For BERT Emotions, which predicts from a 13-label taxonomy, exact-match accuracy cannot be computed at the 37-label or 14-category levels due to taxonomy mismatch. At the Ekman 7-category and sentiment 3-category levels, we map its 13 labels to the same targets using shared label names and valence/arousal alignment, then apply the 7$\to$3 cascade from \citet{gong2024mapping}. Table~\ref{tab:multi_level} reports evaluation on the 493 utterances with explicit human emotion annotations.

\begin{table}[t]
\centering
\small
\setlength{\tabcolsep}{4pt}
\begin{tabular}{llcccc}
\toprule
\textbf{Model} & & \textbf{37} & \textbf{14} & \textbf{7} & \textbf{3} \\
\midrule
\multirow{2}{*}{Random baseline}
  & Acc. & 0.035 & 0.209 & 0.270 & 0.582 \\
  & Sim. & 0.947 & 0.951 & 0.947 & 0.968 \\
\midrule
\multirow{2}{*}{BERT Emotions}
  & Acc. & ---   & ---   & 0.335 & 0.582 \\
  & Sim. & 0.947 & ---   & 0.940 & 0.971 \\
\midrule
\multirow{2}{*}{MentalBERT}
  & Acc. & 0.146 & 0.371 & 0.416 & 0.688 \\
  & Sim. & 0.960 & 0.959 & 0.959 & 0.973 \\
\midrule
\multirow{2}{*}{Llama-3.1-8B}
  & Acc. & 0.349 & 0.605 & 0.645 & 0.813 \\
  & Sim. & 0.975 & 0.977 & 0.974 & 0.982 \\
\midrule
\multirow{2}{*}{Llama-3.2-3B}
  & Acc. & 0.369 & 0.645 & 0.667 & 0.842 \\
  & Sim. & \textbf{0.974} & \textbf{0.977} & \textbf{0.975} & \textbf{0.983} \\
\midrule
\multirow{2}{*}{Mistral-7B}
  & Acc. & \textbf{0.418} & \textbf{0.676} & \textbf{0.682} & \textbf{0.844} \\
  & Sim. & 0.977 & 0.977 & 0.975 & \textbf{0.983} \\
\bottomrule
\end{tabular}
\caption{Model evaluation ($N{=}493$) across taxonomy levels: 37 (original), 14 (Plutchik fine-grained), 7 (Ekman basic), and 3 (sentiment). BERT Emotions accuracy at levels 37 and 14 is not reported due to taxonomy mismatch; at levels 7 and 3, its 13 labels are mapped to the same targets via shared names and the \citet{gong2024mapping} cascade. Bold indicates best per metric.}
\label{tab:multi_level}
\end{table}

At the Ekman 7-category level, all three generative models exceed 0.64 accuracy, with Mistral-7B reaching 0.682. At the sentiment level (3 categories), accuracy rises above 0.81 for all generative models. BERT Emotions, now comparable at the mapped levels, achieves 0.335 accuracy at Ekman 7 and 0.582 at sentiment 3, substantially below the generative models but above MentalBERT at the 7-category level. This confirms that the majority of prediction errors are confusions between semantically adjacent labels within the same broad category (e.g., \texttt{hopeless} vs.\ \texttt{worthlessness}, both mapping to \emph{sadness}), rather than gross misclassifications across emotional poles. The convergence of semantic similarity across all taxonomy levels (0.974--0.983 for generative models) further confirms that model choice does not materially affect the analysis. MentalBERT, while substantially better than the random baseline at the sentiment level (0.688 vs.\ 0.582), still lags behind all other models at every taxonomy level.
% \sararevision{Appendix~\ref{app:model_val_appendix}: the BERT Emotions versus MentalBERT ranking and the claim that MentalBERT lags behind all models conflict with Table~\ref{tab:multi_level}. Update this interpretation after verifying the scores.}

We further use the \emph{with-context} configuration for all primary analyses. As we demonstrate in \S\ref{sec:role_diff_appendix} and \S\ref{sec:context_comparison_appendix}, the without-context configuration fails to capture sustained emotional states, produces erratic label sequences (change rate 84.2\% vs.\ 63.9\%), and collapses the texter--volunteer role distinction. The with-context configuration sacrifices some single-utterance accuracy but produces temporally coherent emotion trajectories essential for dynamics analysis.

\paragraph{Design Rationale. }
These results motivate our choice of the generative prompt-engineering approach. First, the structured prompt, which supplies explicit definitions for all 37 categories alongside conversational context, enables nuanced distinctions (e.g., \emph{grateful} vs.\ \emph{hopeful}, \emph{anxious} vs.\ \emph{afraid}) that fixed-vocabulary classifiers collapse. Second, the three-stage cascade (\S\ref{sec:emotion_detection_pipeline}) provides a principled fallback that maintains label validity even when free-form generation fails. Third, the \texttt{Reason:} field offers interpretable justifications auditable for clinical plausibility, a property absent from softmax-based classifiers. Finally, the pipeline's advantage is robust across all three generative models, all six evaluation subsets, and both context conditions, suggesting that the prompt design generalizes well across architectures and conversational structures.

\paragraph{Llama-3.2-3B-Instruct Context Ablation. }
\label{sec:model_comparison_appendix}
Table~\ref{tab:context_ablation_detail} compares the Llama-3.2-3B-Instruct pipeline under both context conditions on 493 human-annotated utterances with explicit emotion labels. The without-context configuration achieves higher single-utterance accuracy, as expected since human annotations were produced without conversational context. However, the with-context configuration produces more temporally coherent label sequences needed for dynamics analysis (\S\ref{sec:role_diff_appendix}).

\begin{table}[H]
\centering
\small
\begin{tabular}{lcc}
\toprule
\textbf{Configuration} & Sim. & Acc. \\
\midrule
Llama-3.2-3B-Instruct (no ctx) & 0.974 & 0.369 \\
Llama-3.2-3B-Instruct (w/ ctx) & 0.972 & 0.302 \\
\bottomrule
\end{tabular}
\caption{Llama-3.2-3B-Instruct context ablation ($N{=}493$ utterances with explicit human emotion labels). The without-context configuration scores higher on single-utterance validation but fails to capture temporal dynamics (see \S\ref{sec:context_comparison_appendix}).}
\label{tab:context_ablation_detail}
\end{table}

% \clearpage
\section{Emotion to Polarity Mapping}
\label{sec:polarity_mapping_appendix}

Each of the 37 emotion labels is mapped to a polarity class (negative, neutral, or positive) via the NRC VAD Lexicon. Labels with valence $\leq 0.45$ are classified as negative, valence $\geq 0.55$ as positive, and $0.45 < \text{valence} < 0.55$ as neutral. Table~\ref{tab:polarity_mapping} lists the full mapping.

\begin{table*}[t]
\centering
\small
\begin{tabular}{llrl}
\toprule
\textbf{Emotion Label} & \textbf{VAD Term(s)} & \textbf{Valence} & \textbf{Polarity} \\
\midrule
\multicolumn{4}{l}{\emph{Negative (valence $\leq$ 0.45), 21 labels}} \\
\midrule
worthlessness & worthless, worthlessness & 0.042 & Negative \\
fear         & fear, afraid          & 0.042 & Negative \\
shame        & shame, shameful       & 0.050 & Negative \\
disgust      & disgust, disgusted    & 0.052 & Negative \\
hopeless     & hopeless, hopelessness & 0.060 & Negative \\
distress     & distress, distressed  & 0.108 & Negative \\
disappointment & disappointment      & 0.115 & Negative \\
tired        & tired                 & 0.125 & Negative \\
anger        & anger, angry          & 0.145 & Negative \\
boredom      & bored, boredom        & 0.160 & Negative \\
worry        & worry, worried        & 0.170 & Negative \\
stress       & stress, stressed      & 0.170 & Negative \\
guilt        & guilt, guilty         & 0.172 & Negative \\
loneliness   & lonely, loneliness    & 0.198 & Negative \\
regret       & regret, regretful     & 0.199 & Negative \\
numbness     & numb, numbness        & 0.207 & Negative \\
anxiety      & anxiety, anxious      & 0.214 & Negative \\
sadness      & sad                   & 0.225 & Negative \\
preoccupied  & preoccupied           & 0.265 & Negative \\
overwhelm    & overwhelm, overwhelmed & 0.296 & Negative \\
distraction  & distraction, distracted & 0.298 & Negative \\
\midrule
\multicolumn{4}{l}{\emph{Neutral ($0.45 <$ valence $< 0.55$), 3 labels}} \\
\midrule
neutral      & neutral               & 0.469 & Neutral \\
mood         & mood, moody           & 0.483 & Neutral \\
resilient    & resilient             & 0.542 & Neutral \\
\midrule
\multicolumn{4}{l}{\emph{Positive (valence $\geq$ 0.55), 12 labels}} \\
\midrule
longing      & longing               & 0.604 & Positive \\
anticipation & anticipation          & 0.698 & Positive \\
self         & self                  & 0.704 & Positive \\
serenity     & serenity, serene      & 0.851 & Positive \\
empathy      & empathy, empathetic   & 0.865 & Positive \\
surprise     & surprise              & 0.875 & Positive \\
gratitude    & gratitude, grateful   & 0.922 & Positive \\
trust        & trust, trustworthy    & 0.929 & Positive \\
hopeful      & hopeful               & 0.947 & Positive \\
happiness    & happiness, happy      & 0.976 & Positive \\
joy          & joy                   & 0.980 & Positive \\
love         & love                  & 0.998 & Positive \\
\bottomrule
\end{tabular}
\caption{Mapping of 37 emotion labels to polarity classes via NRC-VAD valence scores. Labels with valence $\leq 0.45$ are negative, $\geq 0.55$ are positive, and intermediate values are neutral. The label \texttt{chaotic} has no direct entry in the NRC-VAD lexicon and is omitted from polarity analysis.}
\label{tab:polarity_mapping}
\end{table*}

% \clearpage
\section{Multi-Level Emotion Taxonomy Mapping}
\label{sec:multi_level_mapping_appendix}

Table~\ref{tab:37_to_14_mapping} shows the mapping from our 37 emotion labels to the 14 Plutchik fine-grained categories, following the method of \citet{gong2024mapping}. Labels sharing the same name are mapped directly; remaining labels are assigned based on valence and arousal similarity. The 14 categories are then cascaded to coarser levels following the paper's validated hierarchy: 14$\to$7 (Ekman) merges \emph{serenity}$\to$\emph{joy}, \emph{annoyance}$\to$\emph{anger}, \emph{boredom}$\to$\emph{disgust}, \emph{distraction}$\to$\emph{sadness}, \emph{interest}$\to$\emph{anticipation}$\to$\emph{neutral}, and \emph{trust}$\to$\emph{joy}; 7$\to$3 (sentiment) maps \emph{joy}$\to$\emph{positive}, \emph{anger/disgust/sadness/fear/surprise}$\to$\emph{negative}, and \emph{neutral}$\to$\emph{neutral}.

\begin{table}[H]
\centering
\small
\begin{tabular}{ll}
\toprule
\textbf{14-Category Target} & \textbf{Our 37 Labels} \\
\midrule
joy          & joy, happiness, love, gratitude \\
serenity     & serenity \\
trust        & trust, empathy, resilient \\
anticipation & anticipation, hopeful \\
interest     & preoccupied \\
anger        & anger \\
annoyance    & disapointment \\
disgust      & disgust, shame \\
boredom      & boredom \\
sadness      & sadness, hopeless, worthlessness, \\
             & loneliness, numbness, tired, \\
             & longing, regret, guilt \\
fear         & fear, anxiety, worry, stress, \\
             & overwhelm, distress, chaotic \\
surprise     & surprise \\
distraction  & distraction \\
neutral      & neutral, mood, self \\
\bottomrule
\end{tabular}
\caption{Mapping of 37 emotion labels to 14 Plutchik fine-grained categories following \citet{gong2024mapping}. Direct name matches are mapped as-is; remaining labels are assigned via valence/arousal similarity.}
\label{tab:37_to_14_mapping}
\end{table}

\section{Archetype Analysis Details}
\label{app:archetype}
\paragraph{Trajectory Representation. }
For each conversation, we extract the texter's turn-level distress sequence using the emotion scoring scheme, discarding conversations with fewer than four labelled texter turns. Because conversations vary in length, we linearly interpolate each sequence onto a shared grid of $T=20$ equally spaced points on the normalised progress axis $[0, 1]$. This representation preserves trajectory shape while abstracting away absolute duration, enabling direct comparison across corpora.

\paragraph{Pooled clustering.}
We concatenate trajectories from all three sources---zero-shot synthetic, dual-agent synthetic, and CTL---and fit $K$-means with $K=5$ clusters (\texttt{n\_init}$=20$, seed $42$). 
%Nick
Pooling ensures that archetype labels are shared across sources, so differences in archetype prevalence reflect genuine compositional shifts rather than source-specific partitioning artifacts.

\paragraph{Archetype Labeling.}
We characterize each cluster centroid using four shape features: mean distress level, total decline in distress from start to end, the fraction of that decline occurring in the final quarter, and volatility, measured as the standard deviation of first-order differences. Archetype names are assigned directly from these features rather than selected from a fixed candidate list.

Centroids with a total distress decline below $0.20$ are labeled \emph{Unresolved High Distress} when their mean distress is at least $0.65$, and \emph{Persistent Moderate Distress} otherwise. For centroids showing a larger decline, labels depend on its timing: \emph{Late Recovery} when at least $60\%$ of the decline occurs in the final quarter, \emph{Early Resolution} when at most $30\%$ occurs there, and \emph{Steady De-escalation} otherwise. These labels receive the suffix \emph{(high distress)} when mean distress is at least $0.65$. We additionally reserve \emph{Rupture-Repair} and \emph{Oscillatory Distress} for unusually volatile centroids, defined as having volatility above $0.05$ and at least $0.05$ greater than the median volatility of the remaining centroids. No centroid meets this relative-volatility criterion in our data, so these labels are not assigned. When multiple centroids receive the same archetype label, we distinguish them as \emph{higher distress} and \emph{lower distress} according to their mean levels.

\paragraph{Cross-source comparison.}
For each source, we compute the proportion of conversations assigned to each archetype. Treating the CTL distribution as a reference, we quantify how strongly each synthetic condition over- or under-represents particular recovery patterns relative to real crisis conversations (Figure~\ref{fig:archetype}).

\paragraph{Inter-conversation Diversity.}
To test whether synthetic conversations are less varied than real CTL conversations, we measure inter-conversation diversity at the trajectory level. For each source (synthetic zero-shot, synthetic dual-agent, real CTL), we restrict to with-context-labeled conversations, retain texter turns only, and use the distress score. Conversations with fewer than four valid distress values are discarded. Each remaining conversation is summarized as a length-normalized trajectory by linearly interpolating its distress sequence onto a shared grid of $T = 20$ equally spaced points on $[0, 1]$, yielding one 20-dimensional vector per conversation. To keep the three sources on comparable footing, we subsample each source to at most $N = 1000$ conversations (seed $42$) and compute the full pairwise Euclidean distance matrix between trajectories within each source. We summarize inter-conversation diversity as the mean of the off-diagonal entries of this matrix; a higher mean indicates that conversations within the corpus are more dissimilar from one another. Uncertainty is quantified by a nonparametric bootstrap (1{,}000 resamples) over conversation indices, and we test whether CTL is significantly more diverse than each synthetic source by bootstrapping the difference of means and reporting the 95\% confidence interval.

\section{Additional Results}
\subsection{Speaker Role Differentiation}
\label{sec:role_diff_appendix}
With conversational context, texter and volunteer emotion profiles are sharply distinguished along complementary axes. Texter utterances tend to remain in negative states, with negative-polarity persistence of 84.1\%, whereas volunteer utterances more consistently remain in positive states, with positive-polarity persistence of 72.4\%; the two roles therefore exhibit distinct emotion-label profiles. The volunteer's most frequent cross-emotion transition is \texttt{hopeless$\to$hopeful}, consistent with movement from acknowledging distress toward a more hopeful frame. Texter labels are also more variable, with a per-conversation emotion change rate of 62.8\% compared with 52.8\% for volunteers. Texter conversations are additionally highly likely to begin in a negative state, with 90.9\% of first substantive texter emotions labeled negative.
% \sararevision{Appendix~\ref{sec:role_diff_appendix}: update persistence, transition counts, and endpoint percentages wherever they conflict with the main CTL tables and Appendix~\ref{sec:context_comparison_appendix}. Keep all related role comparisons consistent.}

\label{sec:context_ablation}
Without conversational context, this role differentiation largely collapses: persistence gaps narrow (negative: 22.0~pp $\to$ 12.0~pp; positive: 8.9~pp $\to$ 3.1~pp), dominant self-transitions converge on low-specificity labels (\texttt{neutral}, \texttt{sadness}), and both roles end positive at identical rates (37.6\%). These findings confirm that conversational context is essential for the detection model to distinguish speaker roles and recover the therapeutic arc; we therefore use the with-context configuration for all primary analyses. The full with- vs.\ without-context comparison is provided in \S\ref{sec:context_comparison_appendix}.

\subsection{Additional Tables}
Table~\ref{tab:change_rate} reports per-conversation emotion change rates.
\begin{table}[t]
\centering
\small
\begin{tabular}{lccc}
\toprule
\textbf{Role} & \textbf{Mean} & \textbf{Median} & \textbf{SD} \\
\midrule
CTL Texter    & 0.639 & 0.646 & 0.139 \\
CTL Volunteer & 0.560 & 0.559 & 0.151 \\
\bottomrule
\end{tabular}
\caption{Per-conversation emotion change rate for the full CTL analysis corpus (fraction of adjacent utterance pairs with different labels).}
\label{tab:change_rate}
\end{table}

% \clearpage
\section{With- vs.\ Without-Context Comparison}
\label{sec:context_comparison_appendix}
\paragraph{Contexts. } In preliminary study to compare between with- vs. without- context prediction, we test two inference modes on the full CTL analysis corpus:
\begin{itemize}
    \setlength{\itemsep}{0pt}
    \item \textbf{Without context:} The model receives only the target utterance.
    \item \textbf{With context:} The model receives a sliding window of the $N$ most recent utterances (from both speakers) preceding the target, plus the target itself.
\end{itemize}

Table~\ref{tab:context_comparison} presents the full comparison of texter and volunteer emotion dynamics under both context conditions.
% \sararevision{Appendix~\ref{sec:context_comparison_appendix}: this table conflicts with the main CTL tables in utterance counts, polarity persistence, conversation endpoints, and top self-transition counts. Verify the analysis version and filtering, then update all table scores, gaps, and repeated values in the surrounding text. Original values are retained pending verification.}

\begin{table*}[t]
\centering
\small
\begin{tabular}{lcccccc}
\toprule
& \multicolumn{3}{c}{\textbf{With Context}} & \multicolumn{3}{c}{\textbf{Without Context}} \\
\cmidrule(lr){2-4} \cmidrule(lr){5-7}
\textbf{Metric} & \textbf{Texter} & \textbf{Volunteer} & $\boldsymbol{\Delta}$ & \textbf{Texter} & \textbf{Volunteer} & $\boldsymbol{\Delta}$ \\
\midrule
Utterances & 60,247 & 55,771 &  & 60,220 & 55,744 &  \\
\midrule
\multicolumn{7}{l}{\emph{Polarity persistence $P(\text{stay})$}} \\
\quad Negative & 0.848 & 0.628 & 0.220 & 0.684 & 0.564 & 0.120 \\
\quad Neutral  & 0.163 & 0.214 & $-$0.051 & 0.271 & 0.190 & 0.081 \\
\quad Positive & 0.605 & 0.694 & $-$0.089 & 0.318 & 0.349 & $-$0.031 \\
\midrule
\multicolumn{7}{l}{\emph{Change rate (mean $\pm$ SD)}} \\
\quad Emotion (37-label) & 0.639{\scriptsize$\pm$0.139} & 0.560{\scriptsize$\pm$0.151} & 0.079 & 0.842{\scriptsize$\pm$0.104} & 0.868{\scriptsize$\pm$0.088} & $-$0.026 \\
\midrule
\multicolumn{7}{l}{\emph{Conversation arc (\%)}} \\
\quad Start negative & 92.2 & 51.8 & 40.4 & 93.7 & 69.7 & 24.0 \\
\quad End positive   & 63.7 & 61.5 & 2.2  & 37.6 & 37.6 & 0.0 \\
\midrule
\multicolumn{7}{l}{\emph{Top self-transition}} \\
\quad Rank 1 & \multicolumn{2}{c}{hopeless (6,758) / hopeful (14,714)} & & \multicolumn{2}{c}{neutral (2,558) / sadness (2,117)} & \\
\quad Rank 2 & \multicolumn{2}{c}{hopeful (2,791) / hopeless (4,657)} & & \multicolumn{2}{c}{sadness (1,879) / neutral (1,173)} & \\
\bottomrule
\end{tabular}
\caption{Texter vs.\ volunteer emotion dynamics across context conditions. $\Delta$ = Texter $-$ Volunteer. With context, the two roles are sharply differentiated; without context, the distinction largely collapses.}
\label{tab:context_comparison}
\end{table*}

\paragraph{With Context: Clear Differentiation. }
When conversational context is available, the gap in negative persistence between texters and volunteers is 22.0~pp (0.848 vs.\ 0.628), reflecting the fundamental asymmetry between the texter's sustained distress and the counselor's active redirection. The top texter self-transition is \texttt{hopeless$\to$hopeless} (6,758 counts), while for volunteers it is \texttt{hopeful$\to$hopeful} (14,714 counts). The starting polarity distributions also diverge sharply: 92.2\% of texter sequences begin negative versus only 51.8\% of volunteer sequences.

\paragraph{Without Context: Role Collapse. }
Without conversational context, the model treats each utterance as an isolated text fragment, and the texter--volunteer distinction degrades substantially. The negative persistence gap narrows from 22.0~pp to 12.0~pp (0.684 vs.\ 0.564), and the positive persistence gap shrinks from 8.9~pp to 3.1~pp (0.318 vs.\ 0.349). Many counselor messages (e.g., ``That sounds really hard,'' ``I can hear how much pain you're in'') read as negative when stripped of their empathetic, redirective function in the conversational flow.

The change-rate ordering inverts: with context, texter emotion labels change more often than volunteer labels (63.9\% vs.\ 56.0\%); without context, volunteers show a slightly \emph{higher} change rate (86.8\% vs.\ 84.2\%), because topically diverse counselor utterances receive highly variable labels when classified in isolation. The dominant self-transitions converge on low-specificity labels (texter: \texttt{neutral$\to$neutral}, 2,558; volunteer: \texttt{sadness$\to$sadness}, 2,117), and both roles end positive at identical rates (37.6\%), compared to a 2.2~pp gap with context.

Overall, without conversational context, the model treats each utterance in isolation, inflating the emotion change rate (84.2\% vs.\ 63.9\% with context) and collapsing the texter--volunteer role distinction: both roles end positive at identical rates (37.6\%), and the model defaults to \texttt{neutral} for contextually ambiguous utterances such as a volunteer's empathetic reflection of distress. With context, the model can sustain emotional labels across turns, capture the therapeutic arc (92.2\% negative starts shifting to 63.7\% positive endings), and distinguish the complementary roles of texter (anchored in distress) and volunteer (anchored in hope).

\section{Synthetic Generation Conditions}
% \section{Additional Synthetic Dialogue Analyses}
\label{app:synthetic_additional}
% \subsection{Synthetic Generation Conditions}
\label{sec:generation_conditions}

We generate synthetic crisis-support conversations using two complementary strategies. In the \emph{dual-agent} setup, separate LLM-backed agents take asymmetric Texter and Volunteer roles and generate the conversation turn by turn. This paradigm is designed to encourage interactional emergence, allowing each agent to respond to the developing conversation. In the \emph{zero-shot} setup, a single LLM generates a complete conversation in one pass, providing tighter control over the global arc and conversation structure.
No CTL messages, paraphrases, summaries, or other CTL-derived materials were used in either generation setup, including in prompts or seeds.

Across both setups, we use three frontier instruction-tuned models from distinct model families:
\textsc{claude-haiku-4-5-20251001} \cite{anthropic2024claude},
\textsc{gpt-4o-2024-08-06} \cite{openai2024gpt4o}, and
\textsc{Gemini-2.5-pro} \cite{google2025gemini25}.
These models span distinct commercial training pipelines and allow us to examine how model-specific alignment, safety tuning, and stylistic priors affect generated crisis dialogues.

To control topical coverage, we construct a seed inventory spanning two broad classes. Grief-related concerns include parental loss, pet loss, relationship dissolution, job loss, miscarriage, and identity loss. Non-grief concerns include anxiety, depression, academic stress, social anxiety, family conflict, workplace stress, self-esteem, and LGBTQ+ struggles. Each generated texter is conditioned on a sampled seed and prompted to produce an opening message grounded in the assigned situation. The volunteer role is prompted to provide empathetic listening, non-directive questioning, and appropriately paced safety assessment.

For the texter role, we adapt established prompts for LLM-simulated patients \cite{louie-etal-2024-roleplay,louie2025care}. Because no directly analogous prompt exists for the volunteer role, we design volunteer prompts that encode counselor-oriented behaviors while maintaining asymmetric role grounding. 
% Full prompts are provided in \S\ref{app:synthetic_prompts}.

We apply the same \empath{} pipeline to CTL and synthetic conversations, including utterance-level emotion labeling, polarity mapping, transition analysis, distress trajectory construction, volunteer strategy classification, and archetype assignment.

\end{document}